\documentclass[sigconf,nonacm]{acmart}

\setcopyright{cc}

\usepackage{listings}
\usepackage{xcolor}
\usepackage[dvipsnames]{xcolor}
\usepackage{pifont}
\usepackage{booktabs}
\usepackage{wasysym}
\usepackage{circledsteps}
\usepackage{orcidlink}
\usepackage{colortbl}

\definecolor{delim}{RGB}{20,105,176}
\definecolor{numb}{RGB}{106, 109, 32}
\definecolor{string}{rgb}{0,0,0}
\definecolor{celadon}{rgb}{0.67, 0.88, 0.69}

\lstdefinelanguage{json}{
    frame=single,
    numbers=left,
    numberstyle=\tiny\color{gray},
    stepnumber=1,
    rulecolor=\color{black},
    tabsize=2,
    showspaces=false,
    showtabs=false,
    showstringspaces=false,
    breaklines=true,
    postbreak=\raisebox{0ex}[0ex][0ex]{\ensuremath{\color{gray}\hookrightarrow\space}},
    breakatwhitespace=true,
    basicstyle=\ttfamily\tiny,
    upquote=true,
    morestring=[b]",
    stringstyle=\color{string},
    literate=
     *{0}{{{\color{numb}0}}}{1}
      {1}{{{\color{numb}1}}}{1}
      {2}{{{\color{numb}2}}}{1}
      {3}{{{\color{numb}3}}}{1}
      {4}{{{\color{numb}4}}}{1}
      {5}{{{\color{numb}5}}}{1}
      {6}{{{\color{numb}6}}}{1}
      {7}{{{\color{numb}7}}}{1}
      {8}{{{\color{numb}8}}}{1}
      {9}{{{\color{numb}9}}}{1}
      {\{}{{{\color{delim}{\{}}}}{1}
      {\}}{{{\color{delim}{\}}}}}{1}
      {[}{{{\color{delim}{[}}}}{1}
      {]}{{{\color{delim}{]}}}}{1},
}

\lstdefinelanguage{Markdown}{
    basicstyle=\ttfamily\tiny,
    breaklines=true,
    columns=fullflexible,
    upquote=true,
    sensitive=true,
    morecomment=[l]{\#},
    morecomment=[l]{>},
    morestring=[b]",
    morestring=[b]',
    moredelim=[s][\bfseries]{**}{**},
    moredelim=[s][\bfseries\itshape]{_}{_},
    moredelim=[s][\color{blue}]{`}{`},
}

\begin{document}


\title[Word-level Annotation of GDPR Transparency Compliance]{Word-level Annotation of GDPR Transparency Compliance in Privacy Policies using Large Language Models}




 \author{Thomas Cory}
 \orcid{0000-0002-3452-9944}
 \affiliation{%
   \institution{Technische Universität Berlin}
   \city{Berlin}
   \country{Germany}}
 \email{cory@tu-berlin.de}

 \author{Wolf Rieder}
 \orcid{0009-0001-4932-9814}
 \affiliation{%
   \institution{Technische Universität Berlin}
   \city{Berlin}
   \country{Germany}}
 \email{w.rieder@tu-berlin.de}

 \author{Julia Krämer}
 \orcid{0000-0001-8605-90780}
 \affiliation{%
   \institution{Erasmus University Rotterdam}
   \city{Rotterdam}
   \country{Netherlands}}
 \email{j.k.kramer@law.eur.nl}

 \author{Philip Raschke}
 \orcid{0000-0002-6738-7137}
 \affiliation{%
   \institution{Technische Universität Berlin}
   \city{Berlin}
   \country{Germany}}
 \email{philip.raschke@tu-berlin.de}

 \author{Patrick Herbke}
 \orcid{0000-0001-9649-2975}
 \affiliation{%
   \institution{Technische Universität Berlin}
   \city{Berlin}
   \country{Germany}}
 \email{p.herbke@tu-berlin.de}

 \author{Axel Küpper}
 \orcid{0000-0002-4356-5613}
 \affiliation{%
   \institution{Technische Universität Berlin}
   \city{Berlin}
   \country{Germany}}
 \email{axel.kuepper@tu-berlin.de}


\renewcommand{\shortauthors}{Cory et al.}

\begin{abstract}
Ensuring transparency of data practices related to personal information is a core requirement of the General Data Protection Regulation (GDPR). However, large-scale compliance assessment remains challenging due to the complexity and diversity of privacy policy language. Manual audits are labour-intensive and inconsistent, while current automated methods often lack the granularity required to capture nuanced transparency disclosures.

In this paper, we present a modular large language model (LLM)-based pipeline for fine-grained word-level annotation of privacy policies with respect to GDPR transparency requirements. Our approach integrates LLM-driven annotation with passage-level classification, retrieval-augmented generation, and a self-correction mechanism to deliver scalable, context-aware annotations across 21 GDPR-derived transparency requirements. To support empirical evaluation, we compile a corpus of 703,791 English-language privacy policies and generate a ground-truth sample of 200 manually annotated policies based on a comprehensive, GDPR-aligned annotation scheme.

We propose a two-tiered evaluation methodology capturing both passage-level classification and span-level annotation quality and conduct a comparative analysis of seven state-of-the-art LLMs on two annotation schemes, including the widely used OPP-115 dataset. The results of our evaluation show that decomposing the annotation task and integrating targeted retrieval and classification components significantly improve annotation accuracy, particularly for well-structured requirements. Our work provides new empirical resources and methodological foundations for advancing automated transparency compliance assessment at scale.
\end{abstract}


\maketitle

\section{Introduction}
\label{sec:introduction}

The European Union's General Data Protection Regulation (GDPR), enacted in 2018, sets a global benchmark for data privacy and transparency. Among its key provisions, Articles 13 and 14 establish stringent requirements for organisations to provide clear and comprehensive information about data collection, processing, and storage practices~\cite{gdpr_13_14}. 

Privacy policies, written in natural language, have emerged as the de facto standard for informing users about data practices and continue to persist as the primary means of communicating privacy-related information~\cite{morel2020sok}. However, privacy policies are often long and complex~\cite{adhikari2023evolution}, use ambiguous language~\cite{reidenberg2016ambiguity}, contain contradictions~\cite{okoyomon2019ridiculousness}, or omit critical details~\cite{wu2023detection}. These factors, coupled with the use of legal jargon, often make privacy policies difficult to understand for non-experts and hinder verification by regulators, making manual audits resource-intensive and prone to inconsistencies.

Prior research has highlighted the difficulty of manual compliance assessments and the need for scalable, automated solutions capable of systematically analysing privacy policies and identifying areas of potential non-compliance~\cite{wilson-2016}. In response, studies have explored the use of machine learning (ML)~\cite{liu-2018} and natural language processing (NLP) techniques~\cite{bui_automated_2021} to automate the analysis of privacy policies. However, existing approaches continue to face challenges in accurately capturing nuanced legal language and context, particularly regarding the annotation (i.e. identification and labelling) of relevant words and phrases within sentences.

In this work, we address these challenges by introducing a modular, large language model (LLM)-based pipeline for fine-grained, word-level annotation of privacy policies with respect to GDPR transparency requirements. Our approach leverages recent advances in LLMs, passage-level classification, and retrieval-augmented generation (RAG) to deliver context-aware, high-granularity annotations suitable for compliance analysis. The key contributions of this paper are as follows:

\begin{enumerate}
    \item We define an annotation scheme comprising 21 distinct transparency requirements derived from GDPR Articles 13 and 14, ensuring comprehensive coverage of regulatory obligations.
    \item We propose a configurable, multi-stage LLM-based annotation pipeline that integrates upstream passage-level classifiers and RAG, enabling scalable and precise word-level annotation of addressed GDPR transparency requirements.
    \item We introduce a two-tiered evaluation methodology that captures both passage-level classification and word-level annotation quality, and apply this framework in a comparative analysis of seven prominent LLMs, providing insights into their performance across annotation tasks.
    \item To support robust empirical evaluation, we implement an automated pipeline for crawling and preprocessing Android app privacy policies, resulting in a large corpus of 703,791 English-language policies and an evaluation dataset of 200 policies with manually curated word-level annotations adhering to the annotation scheme described above.
    \item We evaluate the proposed annotation pipeline on this evaluation dataset as well as the widely-used OPP-115 dataset, highlighting the applicability and generalisability of our approach across different annotation standards.
\end{enumerate}

\section{The GDPR and Transparency}
\label{sec:gdpr_transparency}

The GDPR establishes a legal framework for data protection and privacy in the European Union, setting stringent requirements for entities processing personal data. A core principle of the GDPR is \textit{transparency}, which mandates that individuals be fully informed about how their data is collected, processed, and shared. To enforce this principle, Articles 13 and 14 GDPR define a structured set of information that data controllers must provide to data subjects. Table~\ref{tab:transparency_requirements} summarises these transparency requirements into a comprehensive annotation scheme, categorising them according to their legal basis within the regulation.

A data controller, as defined in Art. 4(7) GDPR, is the entity that determines the purposes and means of processing personal data and is thus responsible for ensuring compliance. This is particularly relevant for digital applications that process user data, where the applications' developers, who are typically also the controllers, must disclose key details to users under Art. 13 GDPR. Accordingly, transparency requirements (Reqs.) 1 and 2 of the proposed annotation scheme address the provision of the controller’s name and contact details, as well as those of their designated Data Protection Officer (DPO), where applicable (Art. 37 GDPR).

Beyond contact details, controllers must clearly state the purpose of data processing (Req. 5), as required by Art. 13(1)(c) GDPR. This obligation stems from the principle of purpose limitation (Art. 5(1)(b) GDPR), which dictates that data processing must be limited to explicitly stated, legitimate purposes. Furthermore, controllers must disclose the legal basis for processing (Req. 6), selecting from the grounds listed in Art. 6 GDPR. Common justifications in digital applications include user consent, performance of a contract, and legitimate interest~\cite{enisaPrivacyDataProtection2017}. For instance, third-party tracking may only be conducted based on user consent~\cite{kollnig2021fait}. If consent is used as a legal basis, Art. 13(2)(c) GDPR further requires that users be informed about their right to withdraw consent in a manner as simple as giving it (Req. 20). Similarly, if data processing is based on the performance of a contract (Req. 12) or legitimate interest (Req. 7), additional contextual information must be provided.

\begin{table}
\caption{Transparency Requirements derived from GDPR Articles 13 and 14.}
\label{tab:transparency_requirements}
\footnotesize
\begin{tabular}{rlll}
\toprule
 & \textbf{Transparency Requirement} & \textbf{GDPR References} \\
\midrule
1. & Controller Name & 13(1)(a), 14(1)(a) \\
2. & Controller Contact & 13(1)(a), 14(1)(a) \\
3. & DPO Contact & 13(1)(b), 14(1)(b) \\
4. & Data Categories & 14(1)(d) \\
5. & Processing Purpose & 13(1)(c), 14(1)(c) \\
6. & Legal Basis for Processing & 13(1)(c), 14(1)(c) \\
7. & Legitimate Interests for Processing & 13(1)(d) \\
8. & Source of Data & 14(2)(f) \\
9. & Data Retention Period & 13(2)(a), 14(2)(a) \\
10. & Data Recipients & 13(1)(e), 14(1)(e) \\
11. & Third-country Transfers & 13(1)(f), 14(1)(f) \\
12. & Mandatory Data Disclosure & 13(2)(e) \\
13. & Automated Decision-Making & 13(2)(f), 14(2)(f) \\
14. & Right to Access & 13(2)(b), 14(2)(c) \\
15. & Right to Rectification & 13(2)(b), 14(2)(c) \\
16. & Right to Erasure & 13(2)(b), 14(2)(c) \\
17. & Right to Restrict & 13(2)(b), 14(2)(c) \\
18. & Right to Object & 13(2)(b), 14(2)(c) \\
19. & Right to Portability & 13(2)(b), 14(2)(c) \\
20. & Right to Withdraw Consent & 13(2)(c), 14(2)(d) \\
21. & Right to Lodge Complaint & 13(2)(d), 14(2)(e) \\
\bottomrule
\end{tabular}
\end{table}

One of the most significant transparency-related challenges arises in the disclosure of data recipients (Req. 10) under Art. 13(1)(e) GDPR. This requirement is particularly relevant in digital applications, where data is frequently transmitted via embedded third-party software libraries~\cite{binnsThirdPartyTracking2018} that can facilitate extensive data sharing, often outside the direct control of app developers~\cite{kochImpactDefaultMobile2025}. 

Additionally, if data is transferred outside the European Economic Area, controllers must inform users about such third-country transfers (Req. 11). Art. 13(2)(f) GDPR further requires the disclosure of automated decision-making processes (Req. 13), particularly those with legal or similarly significant effects (Art. 22 GDPR).

Users must also be informed about their rights under the GDPR (Reqs. 14–21), which include the right to access, rectification, erasure, restriction, objection, portability, and complaint submission (Arts. 15–21 GDPR). Merely listing these rights is insufficient: controllers must explain their implications and provide practical steps for exercising them~\cite{art29wpGuidelinesTransparencyRegulation2018}.

While Art. 13 GDPR applies when data is collected directly from the data subject, Art. 14 GDPR imposes additional transparency obligations when data is obtained from third-party sources. The primary distinction between these provisions lies in Reqs. 4 and 8: controllers must disclose the categories of data collected and the source of the data, respectively. These requirements mitigate the risk of users being unaware of processing activities that involve their personal data. In the context of digital applications, where data is typically collected directly from users~\cite{nguyenFreelyGivenConsent2022}, Art. 13 GDPR is generally the more relevant provision. However, this distinction is secondary, since Art. 13(1)(c) GDPR mandates controllers to describe the purposes and legal basis, which requires the description of the data (and the categories) processed~\cite{zanfir-fortunaArticle13Information2020}.

Failure to comply with transparency obligations can result in administrative fines of up to 4\% of a company’s annual turnover, as stipulated in Art. 83(5) GDPR. This further underscores the legal and financial significance of ensuring full compliance and the need for robust auditing mechanisms, not only to assist users and regulators, but also the data controllers themselves.

\begin{table*}[ht]
\centering
\tiny
\caption{Summary of Related Work for Automated Privacy Policy Information Disclosure Annotation.}
\begin{tabular}{rllllllcccc}
\toprule
\textbf{Year} & \textbf{Paper} & \textbf{Models} & \textbf{Annotation-level} & \textbf{Annotation Basis} & \textbf{Dataset} & \textbf{\#Privacy Policies} & \multicolumn{4}{c}{\textbf{Evaluation Metrics}} \\
\cmidrule(lr){8-11}
              &                &                 &                           &                           &                &                           & \textbf{A} & \textbf{P} & \textbf{R} & \textbf{F1} \\
\midrule
2014 & Zimmeck and Bellovin~\cite{zimmeck-2014} & NB, Rule classifier 
     & Segment 
     & Privacy Practices\textsuperscript{$\dagger$} 
     & Web Crawl\textsuperscript{$\star$} (Alexa)
     & 150
     & \ding{51} & \ding{51} & \ding{51} & \ding{51} \\
\midrule
2016 & Wilson et al.~\cite{wilson-2016} & SVM, LR, HMM 
     & Manual: Phrase,  
     & OPP schema\textsuperscript{$\dagger$}  
     & OPP-115\textsuperscript{$\star$}  
     & 115
     & \ding{55} & \ding{51} & \ding{51} & \ding{51} \\
     &                              & 
     & Automated: Segment      &          & 
     & 
     &          &          &          &  \\
\midrule
2018 & Liu et al.~\cite{liu-2018} & CNN, SVM, LR 
     & Segment, Sentence 
     & OPP schema\textsuperscript{$\dagger$}  
     & OPP-115
     & 115
     & \ding{55} & \ding{51} & \ding{51} & \ding{51} \\
\midrule
     & Wilson et al.~\cite{wilson-2018} & SVM, CNN, LR 
     & Segment, Sentence 
     & OPP schema\textsuperscript{$\dagger$}  
     & OPP-115
     & 115 
     & \ding{55} & \ding{51} & \ding{51} & \ding{51} \\
\midrule
     & Harkous et al.~\cite{harkous_polisis_2018} & CNN (Hierarchy of classifiers)
     & Segment 
     & OPP schema\textsuperscript{$\dagger$}  
     & Web Crawl\textsuperscript{$\star$}, OPP-115 
     & 130K, 115
     & \ding{51} & \ding{55} & \ding{55} & \ding{55} \\
\midrule
     & Tesfay et al.~\cite{tesfay-2018} & NB, SVM, DT, RF 
     & Sentence 
     & GDPR Privacy Aspects\textsuperscript{$\ast$} 
     & Web Crawl\textsuperscript{$\star$}  (Alexa)
     & 10 
     & \ding{55} & \ding{51} & \ding{51} & \ding{51} \\
\midrule
2019 & Andow et al.~\cite{andow-2019} & NER and heuristics 
     & Sentence 
     & Contradiction types\textsuperscript{$\ast$} 
     & Google Play Store\textsuperscript{$\star$} 
     & 11,430
     & \ding{55} & \ding{51} & \ding{51} & \ding{55} \\
\midrule
     & Story et al.~\cite{story-2019} & SVC 
     & Sentence 
     & Custom mobile-specific schema\textsuperscript{$\dagger$}  
     & APP-350\textsuperscript{$\star$} 
     & 350 
     & \ding{55} & \ding{51} & \ding{51} & \ding{51} \\
\midrule
2021 & Bui et al.~\cite{bui_automated_2021} & BLSTM-CRF, BERT 
     & Phrase 
     & Refined OPP schema\textsuperscript{$\dagger$} 
     & OPP-115
     & 30 
     & \ding{55} & \ding{51} & \ding{51} & \ding{51} \\
\midrule
     & Thotawaththa et al.~\cite{thotawaththa_automated_2021} & BERT, SVM, NB, BiLSTM 
     & Section 
     & User \& Expert-based categories\textsuperscript{$\dagger$}
     & Web Crawl\textsuperscript{$\star$} (App Store)
     & 1,430 apps 
     & \ding{55} & \ding{51} & \ding{51} & \ding{51} \\
\midrule
     & Alabduljabbar et al.~\cite{alabduljabbar_tldr_2021} & LR, SVM, RF, CNN, DNN, BERT 
     & Segment 
     & OPP schema\textsuperscript{$\dagger$} 
     & Alexa Top-10K, OPP-115
     & 5598, incl. 115
     & \ding{55} & \ding{51} & \ding{51} & \ding{51} \\
\midrule
     & El Hamdani et al.~\cite{hamdani-2021} & XLNet, CNN, T5-11B 
     & Segment 
     & OPP schema\textsuperscript{$\dagger$}, GDPR Art. 13, 14\textsuperscript{$\ast$} 
     & OPP-115 subset, \cite{linden2020privacy} subset
     & 115, 15 
     & \ding{55} & \ding{51} & \ding{51} & \ding{51} \\
\midrule
2022 & Arora et al.~\cite{arora2022tale} 
     & BERT, M-BERT 
     & Phrase 
     & MAPP schema\textsuperscript{$\ast$}  
     & MAPP\textsuperscript{$\star$} 
     & 300 annotated, 
     & \ding{55} & \ding{55} & \ding{55} & \ding{55} \\
     &                              &      
     & 
     & 
     & 
     & 205,973 in total
     &  &  &  &  \\
\midrule
2023 & Xiang et al.~\cite{xiang-2023} & SRL, NER 
     & Sentence 
     & GDPR Art. 13, 14\textsuperscript{$\ast$} 
     & UK Google Play Store\textsuperscript{$\star$} 
     & 300 annotated, 
     & \ding{51} & \ding{51} & \ding{51} & \ding{51} \\
     &                              & 
     & 
     & 
     & 
     & (205,973 in total)
     &  &  &  &  \\
\midrule
     & Tang et al.~\cite{tang_policygpt_2023}\textsuperscript{$\diamond$} & GPT-4, ChatGPT, Claude2, PaLM, Llama 2 
     & Sentence, Segment 
     & GDPR Art. 13\textsuperscript{$\ast$}, OPP-115 schema\textsuperscript{$\dagger$} 
     & PPGDPR, OPP-115
     & 304, 115 
     & \ding{51} & \ding{51} & \ding{51} & \ding{51} \\
\midrule
2024 & Huang et al.~\cite{huang-2024} & GPT-4 Turbo 
     & Phrase 
     & OPP-115 schema 
     & Web Crawl\textsuperscript{$\star$}  (Vanguard, 
     & 2,545 
     & \ding{55} & \ding{51} & \ding{55} & \ding{55} \\
     &                              & 
     & 
     & derivation\textsuperscript{$\dagger$}
     & Russell 3000 ETF)
     & 
     &  &  &  &  \\[1mm]
\midrule
     & Rodriguez et al.~\cite{rodriguez_large_2024} & GPT-4 Turbo, Llama 2 
     & Segment, Paragraph 
     & MAPP-schema\textsuperscript{$\ast$} 
     & OPP-115, MAPP
     & 115, 65 
     & \ding{51} & \ding{51} & \ding{51} & \ding{51} \\
\midrule
2025 & \textbf{Our Approach} & DeepSeek-R1, Gemma-2, GPT-4o Mini, 
     & Word/Phrase 
     & GDPR Art. 13, 14\textsuperscript{$\ast$} 
     & Web Crawl\textsuperscript{$\star$} (Play Store)
     & 200 annotated,
     & $\circ$ & \ding{51} & \ding{51} & \ding{51} \\
     &                   & GPT-4o, Llama-3.3, Mixtral, Phi-4, Qwen-2.5 
     &          &          & 
     & 703,791 in total
     &  &  &  & \\
\bottomrule
\end{tabular}
\captionsetup{justification=centering} 
\caption*{
\footnotesize
$\dagger$ Not GDPR-related (incl. studies where GDPR relation was not identifiable),
$\ast$ GDPR-related, $\diamond$ Preprint, $\star$ Created the dataset\\
A: Accuracy, P: Precision, R: Recall, F1: F1-score, \ding{51} Included, \ding{55} Not included, $\circ$ Not applicable 
}
\label{tab:related_work}
\end{table*}

\section{Related Work}
\label{sec:related_work}

The emergence of privacy policies as the primary medium for the disclosure of privacy-related data practices, coupled with their complexity, has motivated extensive research efforts aimed at automating their analysis.
A persistent challenge in this field is the need for high-quality labelled datasets to serve as ground truth for ML. While expert-annotated datasets are of the highest quality, their creation is resource-intensive and costly. Crowdsourcing has been explored as an alternative~\cite{wilson-2018}, but labour costs remain high, resulting in a reliance on a small number of expert-labelled benchmarks. Among these, OPP-115~\cite{wilson-2016}, a phrase-level annotated corpus of 115 privacy policies introduced in 2016, has become a widely used reference~\cite{liu-2018, alabduljabbar_tldr_2021}.

Despite its prominence, however, OPP-115 is not GDPR-aligned, as it predates the regulation and uses a distinct annotation scheme. More recent efforts have thus focused on extending OPP-115~\cite{arora2022tale} and developing new datasets that explicitly reflect GDPR transparency obligations, particularly those laid out in Articles 13 and 14~\cite{hamdani-2021, xiang-2023, tang_policygpt_2023}. These newer corpora aim for closer alignment with regulatory requirements, but coverage and granularity vary.

Even with the availability of datasets featuring fine-grained manual annotations, such as OPP-115, most automated approaches remain focused on segment- or sentence-level annotation, due in part to model constraints. Early work relied on classical ML models such as Support Vector Machines (SVMs) and Logistic Regression (LR)~\cite{zimmeck-2014, wilson-2016}, later giving way to deep learning models like Convolutional Neural Networks (CNNs)~\cite{liu-2018, harkous_polisis_2018}. The introduction of transformer-based models, including Bidirectional Encoder Representations from Transformers (BERT), enabled more effective word- and phrase-level annotation~\cite{bui_automated_2021, hamdani-2021}, though such methods remain less common. Notably, Xiang et al.~\cite{xiang-2023} applied a combination of NLP methods such as Semantic Role Labelling (SRL) and Named Entity Recognition (NER) to sentence-level annotation schemes reflecting GDPR requirements, but their approach does not extend to phrase-level labelling.

More recent advances have seen LLMs explored for privacy policy analysis. For example, Tang et al.~\cite{tang_policygpt_2023} employed LLMs for segment- and sentence-level classification on the OPP-115 and PPGDPR~\cite{liu2021have} datasets, focusing on high-level categorisation rather than detailed word- or phrase-level annotation. Rodriguez et al.~\cite{rodriguez_large_2024} systematically evaluated hyperparameter and prompt engineering strategies on the MAPP dataset, highlighting LLMs’ sensitivity to configuration but again restricting analysis to segment-level classification.

Table~\ref{tab:related_work} provides an overview of key studies, mapping their methodologies, annotation levels, and datasets. It reflects the field’s progression from classical machine learning to deep learning, trans-former-based models, and more recently, LLMs, alongside an increasing emphasis on GDPR compliance and annotation schemes based on Articles 13 and 14.

Despite these advances, most prior work remains limited to segment- or sentence-level classification, leaving word- and phrase-level annotation underexplored, particularly for the nuanced transparency requirements of GDPR Articles 13 and 14. Even recent studies with regulatory alignment do not offer approaches capable of systematically extracting and labelling individual transparency-related phrases at scale. To date, the application of transformer-based models and LLMs has largely targeted high-level categorisation rather than precise extraction of legal terms and obligations.

Yet, such granularity is essential for downstream applications like compliance auditing, structured data extraction, and automated policy analysis. Segment-level annotation provides only a coarse overview and cannot capture legally significant distinctions within passages. To reliably assess GDPR compliance, it is crucial to accurately identify specific rights, obligations, and data-sharing practices at the word and phrase level.

\section{Research Questions}
\label{sec:research_questions}

Addressing the research gap laid out above, this work explores the potential of LLMs to generate fine-grained annotations of privacy policies that explicitly align with GDPR transparency obligations. Unlike traditional ML approaches that depend on large, manually curated datasets for training, LLMs offer the promise of generating structured annotations with minimal supervision. However, their capacity to perform detailed legal annotation remains underexplored. This leads to two central research questions:

\begin{enumerate}
    \item[\textit{RQ1}] How can LLMs be leveraged to generate fine-grained word-level annotations in privacy policies that align with GDPR transparency requirements?
    \item[\textit{RQ2}] What evaluation methodology is required to assess the reliability, completeness, and regulatory alignment of such fine-grained LLM-generated annotations?
\end{enumerate}

Section~\ref{sec:approach} addresses RQ1 by introducing a structured methodology that integrates a modular annotation approach that integrates LLM-based annotators with passage-level classifiers, RAG, and self-correction mechanisms. 

Addressing RQ2, Section~\ref{sec:evaluation} outlines a two-tiered evaluation framework that assesses annotation quality while accounting for linguistic ambiguity and policy variation. This includes a comparative analysis of seven publicly available LLMs against a manually curated ground truth dataset comprising 200 privacy policies from Android applications.

\section{Approach}
\label{sec:approach}

As outlined above, LLMs demonstrate strong capabilities in natural language understanding, making them a compelling choice for automating the annotation of privacy policies. However, their attention mechanisms~\cite{vaswani2017attention} and finite context windows introduce a trade-off between input length and annotation granularity. While LLMs perform well on broader tasks such as summarisation or classification over longer texts, fine-grained tasks like word- or span-level annotation require a more localised focus. As input length increases, semantic precision may suffer, reducing annotation accuracy.

Therefore, effective annotation requires a balance: too much text risks diluting attention to individual phrases, while overly narrow inputs may omit essential context. This is especially critical for privacy policies, where meaning often hinges on surrounding content.

To address this challenge, our approach combines two stages. First, we convert raw HTML documents into structured lists of textual passages that preserve context while remaining within manageable input sizes. Second, we apply an LLM-based annotation pipeline to analyse each passage, producing GDPR-aligned annotations with word-level precision.

\subsection{Data Preprocessing}
\label{subsec:data_preprocessing}

\begin{figure*}[tb]
  \centering
  \includegraphics[scale=0.9]{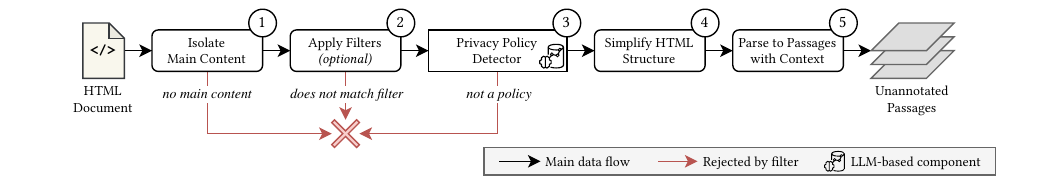}
  \caption{Preprocessing pipeline that parses raw HTML documents into lists of annotation-ready passages.}
  \label{fig:preprocessing_pipeline}
  \Description{Pre-processing Pipeline.}
\end{figure*}

Effective large-scale annotation requires more than managing context length, as it also depends on clean, well-structured inputs.

Despite initiatives advocating for standardised, machine-readable formats~\cite{cranor2003p3p, pandit2019creating}, privacy policies vary significantly in structure and are typically embedded in noisy HTML documents with extraneous elements such as navigation menus, footers, or advertisements. Additional challenges arise due to misconfigured websites, such as broken links redirecting users to generic landing pages rather than the intended privacy policy.

To address these challenges, our preprocessing pipeline transforms raw HTML documents into structured JSON representations optimised for LLM-based annotation. The pipeline, illustrated in Figure~\ref{fig:preprocessing_pipeline}, consists of several steps \Circled{\footnotesize{1}} to \Circled{\footnotesize{5}}, which we describe in the following subsections.

\subsubsection{Isolate Main Content}
\label{subsubsec:isolate_main_content}

We begin by isolating the main content within the given HTML document to ensure that extraneous elements do not interfere with subsequent processing \Circled{\footnotesize{1}}. We adopt a rule-based filtering strategy to detect and remove elements unlikely to contribute meaningful content (e.g., \textit{nav}, \textit{footer}, \textit{script}).

Additionally, a rule-based heuristic identifies the primary content container most likely to hold the privacy policy by examining semantic indicators, such as tag names (e.g., \textit{main}, \textit{article}) and relevant attribute patterns. Documents lacking identifiable policy content (e.g., error pages) are discarded.

\subsubsection{Apply Filters}
\label{subsubsec:apply_filters}

Given the scale and variability of input compiled from automated web crawls, our pre-processing pipeline incorporates a range of customisable filters designed to improve efficiency and reliability by reducing the volume of irrelevant or low-quality documents \Circled{\footnotesize{2}}.

The preprocessing pipeline is explicitly designed to allow researchers to add and configure additional filters tailored to their specific research needs, such as:

\begin{enumerate}
    \item \textit{Language filters:} Ensure that only privacy policies in the desired language(s) are processed. 
    \item \textit{Length-based filters:} Exclude documents that are suspiciously short (e.g., fewer than a few dozen words) or excessively long (e.g., erroneous HTML merges or scripts).
    \item \textit{Deduplication filters:} Detect and remove or group duplicate documents to prevent redundancy and reduce computational overhead.
    \item \textit{Other domain-specific filters:} Apply additional constraints, such as filtering based on keyword presence or specific disclaimers.
\end{enumerate}

Section~\ref{subsubsec:population} describes the filter configuration we used to compile the population from which we draw our evaluation sample.

\subsubsection{Privacy Policy Detector}
\label{subsubsec:privacy_policy_detector}

Whereas prior work employed conventional classification approaches such as Logistic Regression~\cite{zimmeck2019maps} and CNNs~\cite{linden2020privacy} or smaller language models like RoBERTa~\cite{privaSeer} to eliminate non-policy documents, Kostina et al.~\cite{llmForClassification} showed that LLMs can outperform conventional classification models in complex text classification tasks.

Based on this observation, we integrate an LLM-based classifier \Circled{\footnotesize{3}} that analyses a short excerpt ($\sim$200 words) of a given document and assigns one of three labels: \textit{true} if it is a privacy policy, \textit{false} if it is not, or \textit{unknown} if no determination is possible. Documents labelled \textit{false} or \textit{unknown} are excluded. The prompt used for this task is listed in Appendix~\ref{app:prompt_detector}.

On a manually labelled sample of 340 randomly selected documents from our evaluation dataset compilation process (see Section~\ref{subsec:evaluation_dataset}), this approach (using GPT~4.1 as the classifying LLM) achieved an accuracy of 99.7\% with no false negatives and a single false positive, where it incorrectly classified general terms of service as a privacy policy.

\subsubsection{Simplify HTML Structure}
\label{subsubsec:simplify_html_structure}

Next, we prepare the document for passage extraction by simplifying the HTML structure to produce a pseudo-HTML representation that retains only meaningful textual elements while preserving hierarchical cues \Circled{\footnotesize{4}}.

All superfluous elements and attributes, such as scripts or inline CSS styles, are removed. The Document Object Model (DOM) structure is flattened by unwrapping inline elements (e.g., \textit{spans}) and eliminating redundant nesting of elements. Additionally, non-standard tags are converted into \textit{divs} for consistency.

\subsubsection{Parse Passages}
\label{subsubsec:parse_passages}

The simplified HTML structure is parsed into a list of passages $P$, where each element of the simplified DOM is treated as a distinct passage $p \in P$ with text content $v_p$ \Circled{\footnotesize{5}}.

To ensure that each passage retains essential context, we append a list of contextual elements $C_p$ derived from the simplified HTML structure. These contextual elements $c \in C_p$ include relevant section headings, the preceding passage if the passage belongs to a list or table, and, in the case of table cells, the corresponding column and row headers. This approach enables each passage to be examined and annotated separately while preserving a broader contextual understanding, thus addressing the inherent trade-off between input size and context awareness in LLMs.

For further context, we append the passage's HTML element type $e_p$. We distinguish four element types in the simplified DOM:

\begin{enumerate}
    \item \textit{Headlines:} Typically represented by heading tags (e.g., \textit{h1}, \textit{h2}), which delineate sections or subsections.
    \item \textit{Basic text passages:} Single sentences or paragraphs forming the main body of the policy text, denoted by \textit{p} or \textit{div} tags.
    \item \textit{List elements:} Enumerations or bullet lists that organise disclosures about data processing practices, denoted by \textit{li} tags.
    \item \textit{Table cells:} Used in some policies to summarise data collection or usage in tabular format, denoted by \textit{td} tags.
\end{enumerate}

Consequently, the output of this step is a list of all the privacy policy's passages $P$ where each passage $p \in P$ is defined as a 3-tuple:

\begin{equation}
p = (e_p, C_p, v_p) p \in P
\end{equation}

This representation preserves local meaning and broader context while keeping inputs LLM-friendly. The resulting passage list $P$ is passed to the annotation pipeline.

\subsection{Annotation Pipeline}
\label{subsec:annotation_pipeline}

\begin{figure*}[tb]
  \centering
  \includegraphics[scale=0.9]{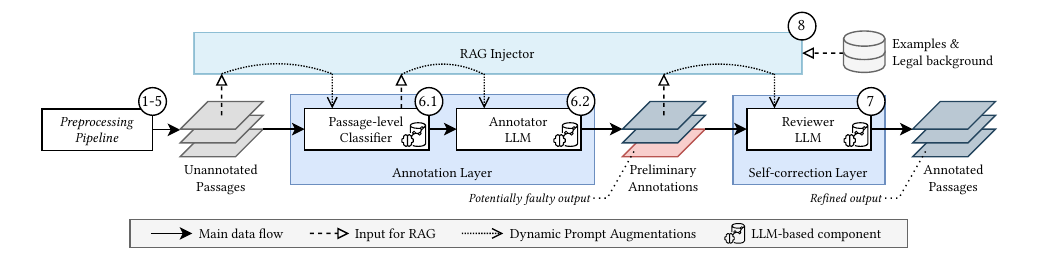}
  \caption{Annotation Pipeline comprising LLM-based annotation- and self-correction layers. The annotation layer combines its core LLM-based annotator with an upstream passage-level classifier to produce annotated passages, which are refined by the self-correction layer's LLM-based reviewer. All LLM-based components are supported by a dedicated RAG injector, which dynamically augments their inputs with suitable examples and legal background.}
  \label{fig:annotation_pipeline_revised}
  \Description{LLM Annotator pipeline.}
\end{figure*}

The annotation pipeline~\footnote{https://github.com/tomcory/privacy-policy-annotator} augments the structured passages from the preprocessing stage with GDPR-aligned, fine-grained annotations. As illustrated in Figure~\ref{fig:annotation_pipeline_revised}, the pipeline comprises two layers: the annotation layer generates initial word-level annotations, and the self-correction layer revises and refines them. This section defines the annotation schema and details the function of both layers.

\subsubsection{Annotations}
\label{subsubsec:annotations}

Within the context of this research, \textit{annotating} refers to the systematic identification and labelling of specific text spans within privacy policies according to predefined categories, namely legally mandated disclosures under GDPR Articles 13 and 14. An \textit{annotation}, therefore, is a labelled span of text relevant to a GDPR transparency requirement. We represent each annotation as a 3-tuple $a_t = (t, l, b)$, where:

\begin{itemize}
    \item $t$ is the relevant text span comprising one or more words (e.g., \textit{"your email address"});
    \item $l$ is the associated label (e.g., \textit{"Data Categories"});
    \item $b \in \{0,1\}$ indicates whether the statement affirms ($1$) (e.g., \textit{"we collect X"}) or denies ($0$) (e.g., \textit{"we do not collect X"}) the addressed requirement.
\end{itemize}

We define specific annotation guidelines to ensure consistency and precision. General introductions, explanatory content, and references to other sections are excluded, as they do not directly reflect GDPR transparency disclosures. Annotations target the smallest relevant text span that clearly expresses the requirement, including necessary restrictive or defining clauses (e.g., \textit{“your name”}, \textit{“other companies we are affiliated with”}). Irrelevant injected clauses are omitted; thus, annotations may span discontinuous segments (e.g., \textit{“your [...] e-mail address”}). Passages may contain multiple relevant phrases and therefore receive multiple annotations, potentially with the same label. Phrases can also fulfil several GDPR requirements simultaneously, leading to multiple annotations with different labels. For example, \textit{“promoting our business through marketing”} may qualify as both a \textit{Processing Purpose} and, if explicitly stated, a \textit{Legitimate Interest for Processing}. In cases of conjunction reduction, where a shared phrase applies to multiple items but is stated only once, each annotation must include the full context. For instance, in \textit{“You have the right to access and delete your data”}, two annotations are required: \textit{“You have the right to access [...] your data”} and \textit{“You have the right to [...] delete your data”}.

\subsubsection{Annotation Layer}
\label{subsubsec:annotation_layer}

The annotation layer processes the set of passages $P$ produced by the preprocessing pipeline presented in Section~\ref{subsec:data_preprocessing}.

Each passage is processed individually by the annotation layer to ensure that the annotator LLM receives input within its optimal context window, minimising unnecessary information while maintaining sufficient context for accurate annotation. The task of the annotation layer is to generate a set of annotations $A_p$ for a given passage by identifying relevant text spans $t \in v_p$ according to the annotation guidelines outlined in Section~\ref{subsubsec:annotations}. 

However, text annotation is inherently a compound task for machine learning models, as it requires the models to identify which labels are addressed by the passage and localise the precise words that correspond to each label. By breaking this compound task into distinct passage-level classification and word-level annotation tasks, we reduce the complexity of each sub-task. Previous research has mainly focused on conventional classifiers for the first sub-task, as exemplified by Xiang et al.~\cite{xiang-2023}, who developed a sentence-level classifier for GDPR requirements with an overall F1-score of 80\%. However, recent work by Kostina et al.~\cite{llmForClassification}, as well as our own findings with the privacy policy detector (Section~\ref{subsubsec:privacy_policy_detector}), demonstrate that LLMs can substantially outperform these conventional approaches on complex classification tasks.

Motivated by these findings, the annotation layer integrates an LLM-based passage-level classifier \Circled{\footnotesize{6.1}}. This LLM-based classifier predicts the set of labels $L_p$ relevant to each passage such that

\begin{equation}
L_p = \{\, l \in \mathcal{L} \mid f(p, l) = 1 \,\}
\end{equation}

where $\mathcal{L}$ is the set of all labels and $f(p, l)$ is an indicator function that returns 1 if the classifier predicts that $l$ is present in $p$.

Given $L_p$, the word-level annotator model \Circled{\footnotesize{6.2}} focuses solely on identifying the exact spans for each predicted label rather than performing full multi-label classification from scratch. This modularisation narrows the search space for annotation with the goal of reducing LLM hallucination and supporting fine-grained span extraction even in complex legal texts.

Each identified span is assigned an annotation $a_t = (t, l, b), a_t \in A_p$, where $l$ denotes the addressed GDPR transparency requirement and $b$ indicates whether the phrase affirms or negates the requirement. This process transforms each passage into an annotated 4-tuple:

\begin{equation}
p' = (e_p, C_p, v_p, A_p), \quad p' \in P'
\end{equation}

where $A_p$ represents the set of generated annotations for the passage and $P'$ is the set of preliminarily annotated passages, which is passed to the subsequent self-correction layer.

\subsubsection{Self-correction Layer}
\label{subsubsec:self_correction_layer}

The self-correction layer refines the preliminary annotations generated by the annotation layer by systematically reviewing the set of annotations $A_p$ for each passage $p' \in P'$. This step is necessary to mitigate potential errors in label assignment, phrase identification, and annotation completeness.

The self-correction layer leverages an LLM-based reviewer \Circled{\footnotesize{7}}, which may perform any number of the following actions to revise $A_p$:

\begin{enumerate}
\item Modify an existing annotation's label $l$ or boolean value $b$ if errors or inconsistencies are detected.
\item Remove an annotation if it is determined to be incorrect or irrelevant based on the context of the passage.
\item Add a new annotation $a_t$ to a text span $t$ that meets the criteria for annotation outlined in Section~\ref{subsubsec:annotations}.
\item Correct an inaccurately identified span (i.e. a span where either too few or too many words are included) by applying (2) to remove the inaccurate annotation and (3) to add a new annotation to the correct span.
\end{enumerate}

Following this review process, each passage's set of annotations $A_p$ is updated to a revised set $ A'_p$, transforming the passage representation into:

\begin{equation}
p'' = (e_p, C_p, v_p, A'_p), \quad p'' \in P''
\end{equation}

where $P''$ is the refined set of annotated passages, which serves as the final output of the annotation pipeline (see Appendix~\ref{app:json_schema_policy}) and forms the basis for downstream evaluation and analysis.

Like all generative ML models, LLMs have a tendency to hallucinate incorrect information~\cite{maynez2020faithfulness}. To mitigate the impact of such hallucinations, we implement guardrails within the annotation and self-correction layers that prevent them from generating annotations that pertain to made-up labels or text spans. Invalid labels are replaced with the valid label with the highest SBERT cosine similarity. If the cosine similarity is too low, the annotation is discarded. Annotations with invalid text spans are discarded outright.

\subsubsection{Prompt Engineering}
\label{subsubsec:prompt_engineering}

When developing an LLM-based annotation system such as the proposed approach, careful attention must be paid to the design and formulation of prompts, as their quality and clarity directly impact the accuracy, consistency, and reliability of the model outputs~\cite{rodriguez_large_2024}.

With this in mind, we adopted an iterative prompt development approach for our annotation pipeline, starting with simple prompts instructing the LLMs to "annotate or review a given passage based on the transparency requirements mandated by GDPR Articles 13 and 14". The prompts then underwent an iterative refinement process based on evaluations of the LLM-generated output on a sample (n=20) of privacy policy passages and resulting tweaks to address observed errors and inaccuracies. Through this feedback loop of manual evaluation and tweaking, the prompts evolved to encompass several components, each serving a distinct purpose:

\begin{enumerate}
    \item \textit{Background and Context}: Briefly outlines the regulatory environment to anchor the LLM and its task within the broader legal context.

    \item \textit{Task Definition}: Explicitly defines the annotation task as identifying and labelling specific phrases within privacy policies according to GDPR transparency requirements. This task definition is expanded for the self-correction layer to clarify that the input comprises passages with pre-existing annotations to be reviewed and corrected.

    \item \textit{Label Classes and Legal References}: Enumerates and describes the 21 GDPR transparency requirements presented in Section~\ref{sec:gdpr_transparency}, including specific references to relevant GDPR paragraphs and concise illustrative examples of relevant phrases for each transparency requirement. If the annotation layer is configured to use a passage-level classifier, the RAG injector modifies the task description accordingly and replaces the full list of transparency requirements with the labels predicted by the classifier.
    
    \item \textit{General Annotation Guidelines}: Provides guidelines outlining which content should and should not be annotated. Initial observations during the iterative prompt development revealed a tendency for LLMs to annotate too eagerly, leading to increased false positives. Thus, the primary objective of these guidelines is to curb this over-eagerness and reduce the incidence of incorrect annotations.
    
    \item \textit{Linguistic and Grammatical Instructions}: Offers explicit instructions addressing the language-specific annotation criteria outlined in Section~\ref{subsubsec:annotations}, such as handling conjunction reductions, restrictive clauses, and interruptions by irrelevant clauses.

    \item \textit{Additional legal context and background}: Provides relevant legal context, such as GDPR excerpts and recitals.
    
    \item \textit{Output Format}: Instructs the model on the necessary output content and provides a JSON schema detailing the exact output format to ensure that the generated annotations are directly usable by subsequent processing steps.
\end{enumerate}

The full prompts used for classification, annotation, and self-correction are provided in Appendix~\ref{app:prompts}.

\subsubsection{RAG Injector}
\label{subsubsec:rag}

RAG is a technique that enhances the performance of LLMs by supplementing their input with relevant external information retrieved from a curated knowledge base. Instead of relying solely on the information encoded in the model’s parameters, RAG dynamically incorporates retrieved content into the model’s prompt, thereby improving factual accuracy, contextual grounding, and task-specific performance~\cite{lewis2020retrieval}.

To support the annotation and self-correction layers of our pipeline, we implement a custom RAG injector \Circled{\footnotesize{8}} designed to augment the LLMs' prompts with supplementary material relevant to the task at hand. Our RAG database consists of two main resource types:
\begin{enumerate}
\item \textit{Legal background}: GDPR excerpts, recitals, and related commentary for each transparency requirement.
\item \textit{Illustrative examples}: Our implementation uses manually labelled and annotated policy passages that illustrate human-approved phrasing and label assignment. The database is constructed from a 10\% sample of our manually curated evaluation dataset (see Section~\ref{subsec:evaluation_dataset}, with the remaining 90\% reserved exclusively for evaluation to prevent data leakage.
\end{enumerate}

Data is stored as structured records linking each example passage to its annotations, associated labels, and corresponding legal references. Embedded representations of the passages are stored in a vector database to support efficient semantic similarity search, leveraging state-of-the-art vector retrieval methods. Our pipeline employs two complementary task-specific retrieval strategies:

\begin{enumerate}
\item \textit{Similarity-based RAG}: For every input, the RAG injector retrieves example passages from the database based solely on semantic similarity to the input passage, without considering label information. This baseline retrieval forms the foundation for prompt augmentation and is applied at all pipeline stages.
\item \textit{Label-aware RAG}: Builds upon similarity-based retrieval by incorporating label-specific retrieval. For each predicted (or assigned) label associated with a passage by the upstream pipeline, the injector fetches examples and legal background material that are specifically relevant to that label.
\end{enumerate}

Applied retrieval strategies differ between pipeline steps. For the passage-level classifier, only similarity-based RAG is applied, as label predictions are not yet available. For the annotator and reviewer, however, both similarity-based and label-aware RAG are used. The injector retrieves semantically similar manually annotated passages from the vector database and fetches additional annotation examples and legal background for each relevant label. The retrieved materials are compiled into an augmented prompt, which is passed to the LLMs and helps align model outputs with annotation requirements and regulatory expectations.

\section{Evaluation}
\label{sec:evaluation}

To assess the performance of our annotation approach, we conduct two experiments using datasets with different annotation schemes:

\begin{itemize}
    \item \textit{Experiment 1} evaluates our approach when applied to the GDPR transparency annotation scheme introduced in Section~\ref{sec:gdpr_transparency}, using a manually curated dataset of 200 annotated privacy policies.
    \item \textit{Experiment 2} evaluates the generalisability of our approach by applying it to the OPP-115 dataset and annotation scheme.
\end{itemize}

In both experiments, we compare different pipeline configurations, ranging from a standalone annotator LLM to full pipeline configurations that include the RAG injector and the annotation layer's passage-level classifier. This ablation setup allows us to quantify the contribution of these supporting components to the overall performance.

This section details the dataset construction, manual review procedure, and our two-tiered evaluation methodology which assesses performance at passage-level and span-level granularity. It also outlines our selection of LLMs used for the evaluation and presents comparative performance results across the selected LLMs.

\subsection{Evaluation Dataset}
\label{subsec:evaluation_dataset}

A robust ground truth is essential for evaluating our annotation approach when applied to the GDPR transparency requirements introduced in Section~\ref{sec:gdpr_transparency}. However, as discussed, existing datasets use annotation schemes that are incompatible in granularity and label structure. To address this, we compile a dataset of privacy policies with manually curated annotations adhering to our annotation scheme. 

Given the evolving nature of privacy policies that is driven by changing regulations and best practices, we adopt a two-stage strategy to construct an up-to-date and representative evaluation dataset. First, we compile a fresh large-scale corpus of privacy policies rather than rely on corpora such as APP-350~\cite{story-2019} or PrivaSeer~\cite{privaSeer} that are generally a few years old. From this fresh corpus, we draw a representative sample that captures diverse writing styles and structural patterns, enabling a broad and realistic assessment of our approach's performance.

\begin{figure}[tb]
  \centering
  \includegraphics[scale=0.9]{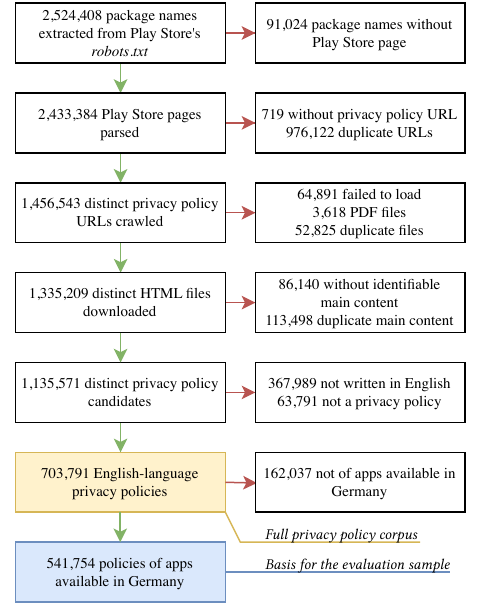}
  \caption{Corpus compilation process, showing the output of each step leading to the corpus of privacy policies used as the population from which we draw our evaluation sample.}
  \label{fig:dataset_compilation}
  \Description{The dataset compilation process. TODO}
\end{figure}

\subsubsection{Population}
\label{subsubsec:population}

When determining the source population for our evaluation sample, the privacy policies of mobile applications are particularly relevant. This is primarily due to the nature of mobile devices, such as smartphones and smartwatches, which are equipped with numerous sensors and operate within data-rich environments. This is coupled with the fact that mobile applications run as native software on devices, granting them greater access to personal and sensitive information from these sources than websites running in sandboxed browser environments~\cite{kollnig2022iphones}.
Consequently, we can reasonably expect their privacy policies to contain more disclosures addressing GDPR transparency requirements~\cite{cory2024mhealth}.

Another major advantage of leveraging mobile applications is the structured availability of privacy policies on app distribution platforms. The iOS App Store and Google Play (also referred to as the Play Store), the predominant digital distribution services within the iOS and Android ecosystems, mandate that developers publish links to their privacy policies on their apps' store pages.

Building on this observation, we developed a custom web crawler to automate the retrieval process of Android app privacy policies. The crawler first extracts a comprehensive list of app identifiers from the Play Store's sitemap. It then crawls each app's Play Store page to extract metadata, including the provided privacy policy URLs. Subsequently, it retrieves and downloads the privacy policies, which are passed to our preprocessing pipeline (see Section~\ref{subsec:data_preprocessing}).

Figure~\ref{fig:dataset_compilation} illustrates the dataset compilation process, initiated in late December 2024. Of the 2,524,408 unique package names extracted from the sitemap, 2,433,384 resolved to active Play Store pages. Parsing these pages resulted in 1,456,543 distinct privacy policy URLs. Following document crawling, deduplication, filtering of non-English documents using \textsc{langid.py}~\cite{langid}, and exclusion of texts not recognised as privacy policies by the LLM-based privacy policy detector (Section~\ref{subsubsec:privacy_policy_detector}), we compiled a corpus of 703,791 distinct English-language privacy policies.

Since our primary objective is to evaluate the annotation of GDPR transparency requirements, we further refine our population to include only privacy policies of applications available for download in Germany, the most populous country in the European Union and the country where the crawl was conducted, yielding 541,754 relevant documents. This final population serves as the basis for our evaluation dataset.

\subsubsection{Sample}
\label{subsubsec:sample}

\begin{figure}[tb]
\centering
\includegraphics[scale=0.8]{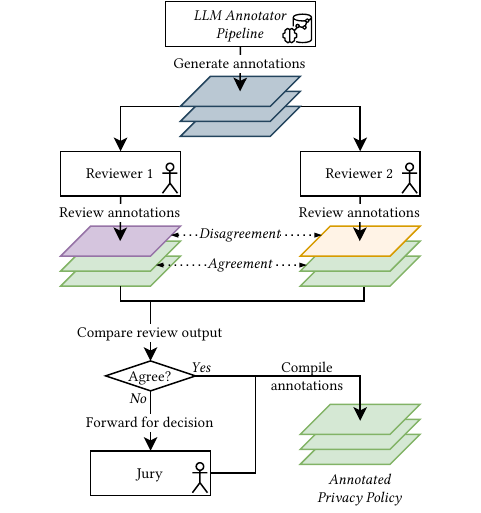}
\caption{Manual review process.}
\label{fig:manual_review_pipeline}
\Description{The manual review process. TODO}
\end{figure}

Having identified a suitable population of privacy policies, we proceed with compiling our evaluation sample.

Privacy policies exhibit substantial variability in length, structure, and language complexity. While some policies concisely address only essential legal requirements, others provide extensive details on data processing practices, legal bases, and user rights. A diverse sample allows us to assess whether our annotation approach effectively handles both concise and verbose documents. However, maximising diversity should not come at the cost of representativeness, as our goal is to infer real-world annotation performance.

To ensure a balanced representation, we apply a diversity-aware sampling strategy that employs k-means clustering to group privacy policies based on their SBERT embeddings. This clustering approach captures structural and linguistic similarities between policies, ensuring that our sample reflects key variations within the dataset. The choice of $k=4$, selected using the Elbow Method, balances granularity and interpretability, segmenting the dataset without overfitting to minor variations.

The resulting clusters exhibit varying sizes, densities, and internal variability. To mitigate over-sampling from small and dense clusters containing highly similar privacy policies, we apply a weighting formula to determine the relative contribution of each cluster to the final sample. The weight $w_c$ for each cluster $c \in C$ is computed as follows:

\begin{equation}
w_c=H_c \times S^2_c \times \frac{n_c}{N}
\end{equation}
\[ w_c \in [0,1], \sum w_c = 1 \]

where $H_c$ denotes the Shannon entropy of cluster $c$, $S^2_c$ represents its variance, $n_c$ is its size, and $N$ is the total population size. This formulation ensures that clusters with higher diversity and larger sizes contribute proportionally more to the sample.

Given a total sample size of $s$, the number of privacy policies selected from each cluster ($s_c$) is determined as:

\begin{equation}
s_c = w_c \times s
\end{equation}

To further enhance intra-cluster diversity, we apply binned stratified sampling, ensuring a balanced representation of diverse document characteristics from each cluster.

We use a sample size of 200 privacy policies to balance practical constraints with statistical robustness, ensure a sufficiently diverse representation of privacy policies, and remain feasible for manual annotation and evaluation. Appendix~\ref{app:sample} summarises the cluster characteristics of our dataset, as well as the computed sample weights and cluster sample sizes.

\subsubsection{Manual Review Process}
\label{subsec:manual_review_process}

To establish a high-quality ground truth for evaluating the LLM-generated annotations, we incorporate a manual review process in which human reviewers systematically revise and validate the annotations. This process mirrors the self-correction layer of the annotation pipeline outlined in Section~\ref{subsec:annotation_pipeline}, but is conducted by domain experts to ensure accuracy and reliability. These reviewers receive detailed briefing sessions and are provided with the same annotation guidelines that were used to instruct the annotation pipeline's LLMs.

To facilitate an efficient and accurate review process, we developed a custom annotation review tool that provides an intuitive graphical interface for human reviewers. The tool highlights annotated phrases in colour-coded formats and allows reviewers to make corrections using simple click-based actions (see Appendix~\ref{app:review_tool}).
This design minimises friction in the review process, allowing reviewers to focus on annotation quality rather than interface complexity.

The manual review workflow is illustrated in Figure~\ref{fig:manual_review_pipeline}. Each LLM-annotated passage $p \in P''$ is passed to two human reviewers, who evaluate it independently to produce two revised annotation sets ${A''_p}^{(1)}$ and ${A''_p}^{(2)}$. The outcomes of these independent reviews are compared and processed on a per-passage basis as follows:

If the revised annotation sets are identical, i.e., ${A''_p}^{(1)} \equiv {A''_p}^{(2)}$, the passage is accepted as final.

If the revised annotation sets differ, i.e., ${A''_p}^{(1)} \not\equiv {A''_p}^{(2)}$, the passage is flagged for further review by a jury of experts.

For disputed passages, the expert jury examines the alternative annotation sets and either: (1) selects one of the two proposed annotation sets as the final version, or (2) further revises the annotations to correct inconsistencies or inaccuracies.

This process results in the transformation of LLM-annotated passages into accepted and jury-revised annotated passages $p_{gt}$:

\begin{equation}
p_{gt} = (e_p, C_p, v_p, A''_p)
\end{equation}

which are compiled into a final annotated privacy policy document $P_{gt}$ that forms the ground truth for the evaluation methodology presented in the following section.

Using this manual review approach, we employed a team of six domain experts from two universities in Germany and the Netherlands to review the LLM-generated annotations for the sample of 200 privacy policies. The policies were equally distributed among the six experts, with two experts randomly assigned to each policy. After being thoroughly briefed on the task and the applied annotation scheme, the experts reviewed their assigned policies independently of each other.

This initial round of independent reviews resulted in a moderate inter-reviewer agreement (Krippendorff's Alpha) of 0.669, highlighting the complexity of this annotation task and the importance of robust mechanisms to resolve disagreements.
Following the approach laid out above, three of the experts formed a jury to jointly examine and discuss all passages with disagreements, either selecting one of the disagreeing options or, in some cases, opting to overrule both initial reviewers.

Through this process, we compiled a set of 25,260 manually curated annotations across the sample of 200 privacy policies to create the \textit{GDPR-Transparency-200} corpus, which serves as the ground truth for our evaluation.

\subsection{Evaluation Methodology}
\label{subsec:evaluation_methodology}

As described at the beginning of this section, we conduct our evaluation using two distinct datasets: the newly created GDPR-Transparency-200 and the OPP-115 dataset. Using the GDPR-Trans-parency-200 dataset, we assess performance under our primary annotation scheme. In contrast, OPP-115 applies a non-GDPR-aligned annotation scheme with different annotation guidelines, thus enabling us to evaluate the generalisability of our pipeline to varied annotation standards and policy corpora.

As our annotation pipeline comprises multiple components, all of which influence the pipeline's output, we perform an ablation study to evaluate the impact of each component individually and in concert with the surrounding components. Specifically, we measure the performance of:

\begin{enumerate}
    \item the standalone annotator with no additional components,
    \item the annotator with downstream self-correction,
    \item the annotator combined with an upstream passage-level classifier and RAG functionality, and
    \item the full pipeline, comprising a passage-level classifier, annotator, and self-correction, all with RAG functionality enabled.
\end{enumerate}

This structured approach allows us to isolate the contributions and interplay of pipeline components systematically and draw conclusions as to which components are especially impactful and which components may benefit from future enhancements.

Given the complexity of our word-level annotation task, traditional multi-label classification metrics are insufficient as they only consider binary presence or absence of labels. Instead, we adopt a two-tiered evaluation framework that measures (1) passage-level performance, examining the passage-level prediction of labels irrespective of annotated text spans, and (2) span-level performance, quantifying the accuracy of annotated text spans associated with each label.

\subsubsection{Level 1: Passage-Level Performance}
\label{subsubsec:label_level_performance}

Passage-level evaluation measures the pipeline’s ability to correctly identify applicable annotation labels within passages. This is formulated as a multi-label classification problem, where we extract predicted label sets $L_p$ from annotations $A_p$ and compare them to ground truth labels $L_p^{gt}$ derived from manual annotations $A''_p$. Both label sets are represented as binary vectors over the complete set of labels, facilitating direct computation of multi-label classification metrics. We calculate \textit{precision}, \textit{recall}, and \textit{F1-score}, and micro-average these metrics across all passages to yield comprehensive performance indicators.

\subsubsection{Level 2: Span-Level Performance}
\label{subsubsec:span_level_performance}

Span-level evaluation assesses how accurately the model identifies the correct text spans for each label. Given that a label is correctly predicted, we evaluate whether the corresponding span(s) align with those in the ground truth.
To quantify this alignment, we compute span similarity using a hybrid metric that combines lexical and semantic similarity:

\begin{equation}
span\theta(t_1, t_2) = \frac{J(t_1, t_2) + cos \theta (t_1, t_2)}{2}
\end{equation}

where $J(t_1, t_2)$ denotes the Jaccard similarity between the two spans, capturing structural similarity by measuring their word overlap, and $cos \theta (t_1, t_2)$ represents the cosine similarity of the SBERT embeddings of the text spans, capturing semantic meaning.

Combining word overlap with semantic embedding similarity ensures that spans are evaluated both on structural alignment and meaning, mitigating potential errors due to minor lexical variations. A purely structural approach, such as Jaccard similarity, only measures direct word overlap and does not account for more nuanced differences. For example, if the model omits a qualifier such as \textit{"your"}, this minor deviation still largely preserves the intended meaning. However, if the model mistakenly includes an unrelated word from an adjacent sentence, this represents a more significant error. Both cases might result in similar Jaccard scores despite differing levels of severity. By incorporating semantic similarity through SBERT cosine similarity, we introduce a finer-grained distinction that allows us to penalise severe mismatches more than minor omissions and ensure that the evaluation more accurately reflects the quality of the model's annotations by distinguishing between small acceptable variations and more substantial errors.

Conversely, the inclusion of structural similarity over an exclusively semantic approach provides robustness against overly broad annotations, i.e., annotations of text spans that go beyond the minimal phrase necessary to convey the required information. If we relied solely on semantic similarity, such undesired additions might not meaningfully change the meaning of the annotated span, resulting in a high semantic similarity score despite the significant structural difference.

Using this similarity function, we assess the quality of the LLM-generated set of annotations $A_p'$ by computing pairwise span similarity with the ground truth annotations $A''_p$. A discrimination threshold $\tau$ is applied to map the calculated similarity score to a binary value, determining whether a predicted annotation correctly corresponds to a ground truth annotation.

A \textit{True Positive} occurs when there exists a corresponding ground truth annotation $a_{gt} \in A''_p$ with text span $t_{gt}$ for a given LLM-generated annotation $a_o \in A_p'$ with text span $t_o$, and vice-versa, such that:

\begin{equation}
TP_o = \{ a_o \mid \exists a_{gt} \in  : \theta(t_{gt}, t_o) > \tau \land l_{gt} = l_o \} \space, \quad TP_{o} \subseteq A'
\end{equation}
\[
TP_{gt} = \{ a_{gt} \mid \exists a_{o} \in  : \theta(t_{gt}, t_o) > \tau \land l_{gt} = l_o \} \space , \quad TP_{gt} \subseteq A''
\]

where $\theta(t_{gt}, t_o)$ denotes the span similarity between $t_{gt}$ and $t_o$, and $l_{gt}$ and $l_o$ are the labels of the annotations $a_{gt}$ and $a_o$, respectively.

A \textit{False Positive} occurs when no matching ground truth annotation exists for a given LLM-generated annotation, such that:

\begin{equation}
FP = \{ a_o \mid \nexists a_{gt} : \theta(t_{gt}, t_o) > \tau \land l_{gt} = l_o \}
\end{equation}

A \textit{False Negative} occurs when a ground-truth annotation is not matched by any LLM-generated annotation, such that:

\begin{equation}
FN = \{ a_{gt} \mid \nexists a_o : \theta(t_{gt}, t_o) > \tau \land l_{gt} = l_o \}
\end{equation}

Unlike conventional classification tasks, the range of \textit{True Negatives} for phrase-level annotations is vast, as any non-annotated span in a passage could be considered a True Negative. Given that the number of possible spans (i.e. combinations of consecutive and non-consecutive words) in a passage $v_p$ grows combinatorially with the passage length, this results in highly skewed metrics when True Negative-based measures such as accuracy are applied.

We account for this by focusing on metrics that do not rely on True Negatives. Specifically, we employ \textit{Precision}, \textit{Recall}, and \textit{$F_1$-score} as the primary evaluation metrics to provide a meaningful assessment of the annotation pipeline's performance. Unlike in conventional binary classification tasks, where the baseline for these metrics is typically 0.5 due to the expected value of random guessing, the vast True Negative space in our setting causes the expected value, and thus the effective baseline, to approach zero as the average passage length increases. In extreme cases, a model could either generate no annotations at all or annotate every possible span in a passage for each label, both of which would result in performance scores approaching zero.

Following the definitions of \textit{True Positives}, \textit{False Positives} and \textit{False Negatives} set out above, micro-average \textit{Precision}, \textit{Recall} and \textit{$F_1$-score} are defined as:

\begin{equation}
Precision = \frac{|TP_{o}|}{|TP_{o}| + |FP|} \quad\quad Recall = \frac{|TP_{gt}|}{|TP_{gt}| + |FN|}
\end{equation}

\[
F_1 = \frac{2}{Precision^{-1} + Recall^{-1}}
\]

\subsection{Model Selection}
\label{subsec:model_selection}

To evaluate the effectiveness of our approach across a diverse range of model architectures and capacities, we selected seven prominent LLMs that represent a cross-section of the LLM landscape as of May 2025. Our selection includes proprietary and open-source models of varying size, as well as different architectural paradigms such as decoder-only transformers and multi-modal models. Specifically, the models used are:

\begin{itemize}
    \item \textit{DeepSeek-R1~(70B)}~\cite{guo2025deepseek}: A mid-size open-source reasoning model built by distilling the full-size DeepSeek-R1 model's capabilities into LLaMA~3.3~(70B).
    \item \textit{Gemma~3~(27B)}~\cite{team2025gemma}: The medium-sized open-source flagship model of the Gemma family, optimised for efficiency and performance across diverse tasks.
    \item \textit{GPT~4.1}~\cite{achiam2023gpt}: The multi-modal flagship of OpenAI's GPT family of proprietary high-performance models.
    \item \textit{GPT~4.1~nano}~\cite{achiam2023gpt}: A more cost-efficient, lightweight variant of GPT~4.1 designed for lower resource demands.
    \item \textit{LLaMA~3.3~(70B)}~\cite{grattafiori2024llama3}: A medium-size open-source model of Meta’s LLaMA family, popular for multilingual and reasoning tasks.
    \item \textit{Phi-4~(14B)}~\cite{abdin2024phi4}: A compact open-source model, commonly used in settings requiring lower inference cost.
    \item \textit{Qwen~3~(32B)}~\cite{yang2025qwen3}: A medium-size open-source model of the Qwen family, comparable in scale and capabilities to Gemma~3.
\end{itemize}

By evaluating our approach across this varied selection, we aim to assess both its robustness to different model types and sizes and its scalability across a range of deployment scenarios.

\subsection{Evaluation Results}
\label{subsec:evaluation_results}

\begin{figure}[tb]
\centering
\includegraphics[width=\linewidth]{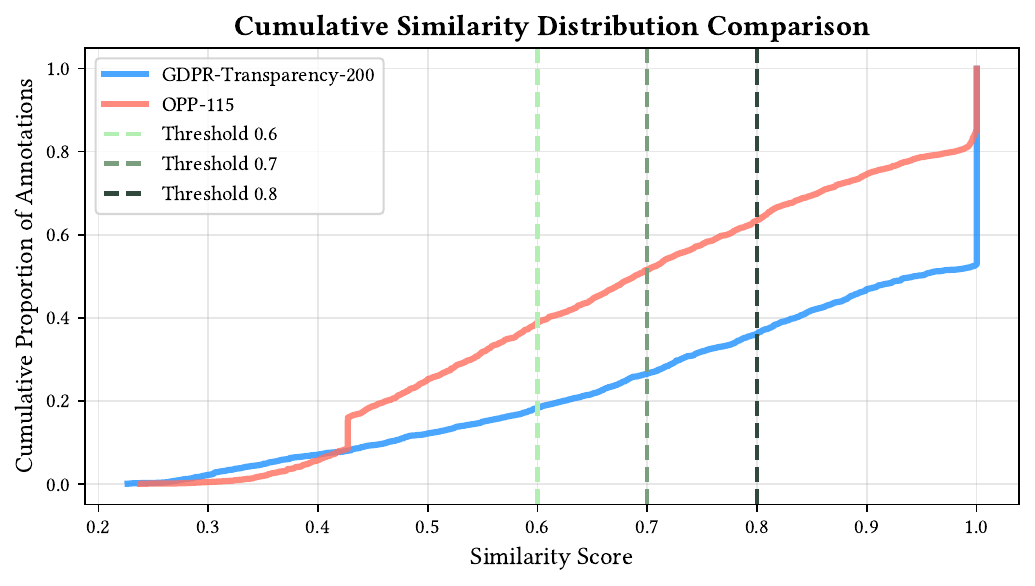}
\caption{Cumulative annotation span similarity distribution, showing the proportion of annotations with span similarity score $span\theta(t_o, t_{gt})$ below given discrimination thresholds $\tau$ across all models and configurations for each dataset.}
\label{fig:combined_cumulative_distribution}
\Description{The manual review process. TODO}
\end{figure}

\begin{table*}[htb]
    \footnotesize
    \caption{Evaluation results for the selected models across datasets and pipeline configurations. Best Precision ($P$), Recall ($R$), and $F_1$-scores per evaluation granularity level, configuration and dataset are highlighted in \textbf{bold}. The best performance scores for each dataset across all configurations are highlighted in \colorbox{celadon}{green}.}
    \label{tab:integrated_results}
    \begin{tabular}{l||ccc|ccc||ccc|ccc}
        \toprule
          & \multicolumn{6}{c||}{\textbf{GDPR-Transparency-200 Dataset}} & \multicolumn{6}{c}{\textbf{OPP-115 Dataset}} \\
        \midrule
         & \multicolumn{3}{c|}{Passage-level} & \multicolumn{3}{c||}{Span-level} & \multicolumn{3}{c|}{Passage-level} & \multicolumn{3}{c}{Span-level} \\
        $\downarrow$ Model $\quad$ Metric $\rightarrow$
        & $P$ & $R$ & $F_1$ & $P$ & $R$ & $F_1$
        & $P$ & $R$ & $F_1$ & $P$ & $R$ & $F_1$ \\
        \midrule
        \multicolumn{13}{c}{\textbf{\textit{(1) Baseline: Standalone annotator (no RAG)}}} \\
        \midrule
        deepseek-r1:70b & 0.719 & 0.579 & 0.641 & 0.568 & 0.656 & 0.609 & 0.740 & 0.699 & 0.719 & \textbf{0.492} & 0.121 & 0.195 \\
        gemma3:27b & 0.659 & 0.779 & 0.714 & 0.544 & 0.758 & 0.634 & \textbf{0.864} & 0.723 & 0.787 & 0.408 & 0.143 & 0.212 \\
        gpt-4.1 & \textbf{0.756} & 0.817 & \textbf{0.785} & \textbf{0.680} & \textbf{0.824} & \textbf{0.746} & 0.839 & 0.810 & \textbf{0.824} & 0.460 & 0.106 & 0.172 \\
        gpt-4.1-nano & 0.496 & 0.659 & 0.566 & 0.414 & 0.583 & 0.484 & 0.768 & 0.709 & 0.737 & 0.379 & 0.098 & 0.156 \\
        llama3.3:70b & 0.616 & \textbf{0.822} & 0.704 & 0.560 & 0.724 & 0.632 & 0.731 & \textbf{0.844} & 0.783 & 0.400 & 0.101 & 0.161 \\
        phi4:14b & 0.626 & 0.662 & 0.643 & 0.502 & 0.670 & 0.574 & 0.666 & 0.772 & 0.715 & 0.397 & 0.092 & 0.150 \\
        qwen3-32b & 0.657 & 0.762 & 0.706 & 0.515 & 0.687 & 0.589 & 0.776 & 0.779 & 0.777 & 0.404 & \textbf{0.161} & \textbf{0.230} \\
        \midrule
        \multicolumn{13}{c}{\textbf{\textit{(2) Standalone annotator with self-correction (no RAG)}}} \\
        \midrule
        deepseek-r1:70b & 0.512 & 0.424 & 0.464 & 0.430 & 0.512 & 0.467 & 0.770 & 0.685 & 0.725 & \textbf{0.504} & 0.095 & 0.160 \\
        gemma3:27b & 0.764 & 0.768 & 0.766 & 0.606 & 0.757 & 0.673 & \textbf{0.868} & 0.730 & 0.793 & 0.414 & 0.133 & \textbf{0.201} \\
        gpt-4.1 & \textbf{0.784} & 0.799 & \textbf{0.791} & \textbf{0.700} & \textbf{0.835} & \textbf{0.761} & 0.850 & 0.803 & \textbf{0.826} & 0.460 & 0.101 & 0.166 \\
        gpt-4.1-nano & 0.460 & 0.430 & 0.444 & 0.390 & 0.564 & 0.461 & 0.771 & 0.709 & 0.739 & 0.381 & 0.095 & 0.152 \\
        llama3.3:70b & 0.570 & \textbf{0.837} & 0.678 & 0.520 & 0.713 & 0.601 & 0.749 & \textbf{0.865} & 0.803 & 0.405 & 0.109 & 0.172 \\
        phi4:14b & 0.611 & 0.559 & 0.584 & 0.493 & 0.611 & 0.546 & 0.691 & 0.782 & 0.734 & 0.430 & 0.098 & 0.160 \\
        qwen3-32b & 0.655 & 0.779 & 0.712 & 0.536 & 0.692 & 0.604 & 0.785 & 0.806 & 0.795 & 0.164 & \textbf{0.153} & 0.159 \\
        \midrule
        \multicolumn{13}{c}{\textbf{\textit{(3) Annotator with upstream passage-level classifier and RAG}}} \\
        \midrule
        deepseek-r1:70b & 0.717 & \textbf{0.758} & 0.737 & 0.687 & 0.637 & 0.661 & 0.843 & 0.685 & 0.756 & \cellcolor{celadon}{\textbf{0.603}} & 0.128 & 0.211 \\
        gemma3:27b & 0.799 & 0.719 & 0.757 & 0.730 & 0.731 & 0.731 & \cellcolor{celadon}{\textbf{0.899}} & 0.737 & 0.810 & 0.519 & 0.164 & 0.249 \\
        gpt-4.1 & 0.856 & 0.748 & \textbf{0.798} & 0.729 & \textbf{0.829} & \textbf{0.776} & 0.876 & 0.806 & \textbf{0.840} & 0.573 & \cellcolor{celadon}{\textbf{0.181}} & \cellcolor{celadon}{\textbf{0.276}} \\
        gpt-4.1-nano & \cellcolor{celadon}{\textbf{0.857}} & 0.514 & 0.643 & 0.693 & 0.609 & 0.648 & 0.861 & 0.536 & 0.661 & 0.544 & 0.129 & 0.208 \\
        llama3.3:70b & 0.613 & 0.399 & 0.483 & \cellcolor{celadon}{\textbf{0.737}} & 0.733 & 0.735 & 0.787 & \textbf{0.830} & 0.808 & 0.539 & 0.160 & 0.246 \\
        phi4:14b & 0.749 & 0.742 & 0.745 & 0.557 & 0.557 & 0.557 & 0.829 & 0.737 & 0.780 & 0.569 & 0.131 & 0.213 \\
        qwen3-32b & 0.698 & 0.770 & 0.732 & 0.728 & 0.736 & 0.732 & 0.815 & 0.775 & 0.794 & 0.537 & 0.161 & 0.248 \\
        \midrule
        \multicolumn{13}{c}{\textbf{\textit{(4) Full pipeline: Annotator with upstream passage-level classifier, downstream self-correction, and RAG}}} \\
        \midrule
        deepseek-r1:70b & 0.560 & 0.819 & 0.665 & 0.605 & 0.615 & 0.610 & 0.785 & 0.758 & 0.771 & 0.520 & 0.113 & 0.186 \\
        gemma3:27b & 0.795 & 0.817 & 0.805 & 0.718 & 0.739 & 0.728 & 0.861 & 0.747 & 0.800 & 0.491 & 0.154 & 0.234 \\
        gpt-4.1 & \textbf{0.827} & 0.834 & \cellcolor{celadon}{\textbf{0.830}} & \textbf{0.724} & \cellcolor{celadon}{\textbf{0.844}} & \cellcolor{celadon}{\textbf{0.780}} & \textbf{0.867} & 0.834 & \cellcolor{celadon}{\textbf{0.850}} & \textbf{0.580} & \textbf{0.159} & \textbf{0.250} \\
        gpt-4.1-nano & 0.700 & 0.650 & 0.674 & 0.689 & 0.616 & 0.650 & 0.810 & 0.633 & 0.711 & 0.544 & 0.129 & 0.208 \\
        llama3.3:70b & 0.583 & \cellcolor{celadon}{\textbf{0.865}} & 0.697 & 0.655 & 0.740 & 0.695 & 0.762 & \cellcolor{celadon}{\textbf{0.844}} & 0.801 & 0.500 & 0.144 & 0.224 \\
        phi4:14b & 0.581 & 0.836 & 0.685 & 0.559 & 0.605 & 0.581 & 0.772 & 0.796 & 0.784 & 0.532 & 0.126 & 0.204 \\
        qwen3-32b & 0.665 & 0.858 & 0.749 & 0.716 & 0.755 & 0.735 & 0.816 & 0.799 & 0.808 & 0.506 & 0.143 & 0.223 \\
        \bottomrule
    \end{tabular}
\end{table*}

We report annotation results for both GDPR-Transparency-200 and OPP-115, analysing passage-level and span-level performance across pipeline configurations and models. For span-level evaluation, we set the discrimination threshold to $\tau = 0.7$, as this value offers a reasonable compromise that allows minor deviations from the ground truth (especially in cases with complex sentence structure, where identifying the exact set of relevant words for an annotation is challenging even for humans) without being overly permissive and artificially inflating performance scores. Figure~\ref{fig:combined_cumulative_distribution} visualises the cumulative distribution of annotation span similarity scores across both datasets, with candidate thresholds indicated. As shown, a threshold of $0.7$ balances the trade-off between overly lax and overly strict evaluation, providing a consistent basis for subsequent span-level analyses on both datasets.

Table~\ref{tab:integrated_results} summarises results for all models and configurations. In the baseline setting using standalone annotators, passage-level classification is generally robust: GPT~4.1 achieves the highest precision (0.756) and $F_1$-score on GDPR-Transparency-200 (0.785), while LLaMA~3.3 delivers the highest recall (0.822). Span-level performance is lower and more varied, with $F_1$-scores ranging from 0.484 to 0.746. On OPP-115, span-level performance is much lower (highest $F_1$-score: qwen3:32b with 0.23), as ground-truth annotation span boundaries are often inconsistent and predicted annotation spans rarely match ground truth spans exactly.

Adding the self-correction layer produces mixed results. Stronger models such as GPT~4.1 benefit modestly, but weaker models often see limited gains or even performance drops, suggesting that self-correction can sometimes reinforce existing errors rather than correct them.

Adding the upstream passage-level classifier and RAG functionality markedly improves the performance. Decomposing the annotation process and providing targeted context through RAG yields higher passage—and span-level scores for most models, especially on GDPR-Transparency-200. On OPP-115, span-level F$_1$ scores remain modest due to high variability in ground-truth annotations, but passage-level classification further improves for most models.

The full pipeline, integrating classifier, annotator, self-correction, and RAG, achieves the best overall performance. GPT~4.1 attains the highest passage- and span-level F$_1$-scores (0.83 and 0.78, respectively) on GDPR-Transparency-200. Improvements from self-correction remain inconsistent for weaker models, however, with three out of seven models showing lower span-level performance that in the configuration without self-correction.

A per-label analysis reveals that models perform best on frequently occurring, short-phrase requirements such as \textit{Controller Name} or \textit{Data Categories}. Performance declines on longer, complex categories (e.g., \textit{Data Retention Period}), and models often confuse closely related labels such as \textit{Processing Purpose} and \textit{Legitimate Interests for Processing}, particularly when policy texts lack clear distinctions. All models are prone to occasional label and text span hallucinations, although this is especially pronounced in smaller models and configurations without RAG.

When comparing datasets, passage-level classification performance is higher for OPP-115, likely due to its simpler annotation label scheme, while span-level evaluation is hindered by inconsistent annotation spans in the ground truth. This effect is clearly visible in Figure~\ref{fig:combined_cumulative_distribution}, where the OPP-115 curve rises steeply, indicating a larger share of low-similarity predictions, whereas GDPR-Transparency-200’s curve is more gradual, reflecting closer correspondence between predicted and reference spans. In contrast, GDPR-Transparency-200's stricter annotation guidelines allow higher measured span-level performance despite a lower passage-level performance resulting from the more complex label scheme. This highlights that annotator models are generally capable of identifying the relevant labels for passages, but have difficulty matching the annotated spans of the ground truth if they do not rigorously adhere to strict guidelines addressing linguistic features.

In summary, splitting the annotation task and careful application of RAG and self-correction yield the strongest results, especially for stronger models. However, precise span annotation and accurate attribution of complex label categories remain challenging.

\section{Discussion}
\label{sec:discussion}

Our evaluation demonstrates that the proposed pipeline for fine-grained annotation of GDPR transparency requirements in privacy policies provides clear advantages, while also highlighting several limitations and avenues for future research.

A central empirical finding is the benefit of decomposing the annotation task into modular components. Introducing a dedicated passage-level classifier and RAG mechanisms for all LLM-based components substantially improved both passage-level and span-level annotation quality, particularly on the GDPR-Transparency-200 dataset. The classifier effectively focuses the annotator on relevant label sets, reducing both under- and over-annotation, while RAG provides targeted examples and contextual information that improve the localisation of relevant spans. Notably, all evaluated LLMs benefitted from this approach, regardless of their size or architecture, though the magnitude of improvement varied.

At the same time, our results reveal important trade-offs and sources of error propagation. The self-correction layer, while beneficial for stronger models, produced inconsistent gains for smaller or weaker LLMs and occasionally compounded errors from earlier pipeline stages. This underscores the need for careful calibration of reviewer prompts and selective use of self-correction, depending on the underlying model’s reliability. Additionally, the pipeline remains sensitive to inaccuracies or ambiguities in classifier predictions and retrieval results, as errors at these stages can propagate and ultimately reduce final annotation quality.

Dataset characteristics also play a critical role in determining pipeline effectiveness. The GDPR-Transparency-200 dataset, with its linguistically precise and fine-grained annotation guidelines, yields higher span-level performance and clearer gains from pipeline modularisation. In contrast, the OPP-115 dataset, with its more heterogeneous annotation spans, proves substantially more challenging for span-level evaluation. This difference, evident both in the cumulative similarity distributions and in performance metrics, suggests that the benefits of automated annotation pipelines are maximised when annotation schemes are rigorous and consistent. Thus, while our approach generalises across differing annotation standards, ground-truth quality and consistency remain critical bottlenecks for reliable automation.

A per-label analysis reveals that LLM-based word-level annotation is most effective for short, well-structured requirements such as \textit{Controller Name} and \textit{Data Categories}, but struggles with labels involving longer, context-dependent spans or ambiguous legal disclosures. Model confusion between related categories (e.g., \textit{Processing Purpose} versus \textit{Legitimate Interests for Processing}) remains a challenge, especially when policy texts lack explicit distinctions. Occasional hallucinations of labels or over-annotation, particularly by smaller models or in the absence of RAG, reinforce the need for robust guardrails and post-processing.

From an application perspective, the pipeline provides a promising foundation for scalable, machine-readable privacy policy annotation. Automated annotation has the potential to support service providers in compliance monitoring, enable regulators to conduct systematic audits, and empower end users through more accurate and structured privacy information. Nevertheless, our results make clear that fully automated solutions are not yet a substitute for expert review, particularly for complex, ambiguous, or rare disclosures. Although it is resource-intensive, the human-in-the-loop review process remains essential for practical deployment in the near future.

Looking ahead, future work should focus on improving the reliability of each pipeline stage, including adaptive retrieval, targeted fine-tuning of models and classifiers, and further optimisation of reviewer workflows. Expanding evaluation to broader datasets, languages, and regulatory frameworks will be critical for establishing generalisability and practical impact. The observed dependence of annotation performance on the quality and consistency of ground-truth datasets also highlights the need for community standards in privacy policy annotation. Furthermore, integrating information sources beyond the privacy policy documents themselves, such as app metadata, presents a promising avenue for improving contextual awareness in automated annotation.

Overall, our findings demonstrate that a modular, classifier- and retrieval-augmented LLM pipeline substantially advances automated privacy policy annotation, especially in settings with rigorous annotation schemes. Realising the full potential of such systems will require ongoing technical refinement and continued integration of human expertise to ensure practical accuracy and reliability.

\section{Conclusion}
\label{sec:conclusion}

This work introduces a modular, LLM-based pipeline for fine-grained, word-level annotation of privacy policy texts with respect to GDPR transparency requirements. Our empirical findings highlight that decomposing the annotation task by integrating a passage-level classifier, retrieval-augmented generation, and an optional self-correction stage substantially improves annotation quality, especially for structured and well-defined disclosures such as controller information and data categories. The benefits of this modular approach are most evident on datasets with precise and consistent annotation guidelines.

However, the evaluation also reveals persistent challenges for automated systems in handling complex, ambiguous, or context-dependent transparency requirements, particularly where disclosure spans are long or loosely structured. While stronger models benefit most from pipeline modularisation and context augmentation, error propagation and dataset variability remain limiting factors, especially for rare or nuanced categories.

Although the proposed approach reduces manual annotation effort and enhances annotation granularity and consistency, expert human oversight is still necessary for quality assurance, particularly in edge cases and for high-stakes compliance scenarios. Future research should focus on refining each pipeline stage, including adaptive retrieval strategies, targeted model fine-tuning, and expanded integration of human-in-the-loop workflows. Further, the development and adoption of more rigorous and standardised annotation schemes will be critical for enabling scalable and generalisable compliance annotation. Ultimately, such advances will support more effective, large-scale analysis of privacy policies and facilitate the development of practical, trustworthy tools for compliance monitoring and user-facing privacy transparency.

\begin{acks}
Funded by the European Union -- NextGenerationEU and the German Federal Ministry of Economics and Climate Protection (BMWK) in the research project "Werk 4.0" and supervised by the project sponsor VDI Technologiezentrum GmbH [grant number 13IK022K].

For some sections of this paper, generative AI-based tools were used to revise the text, improve flow, and fix typos and grammatical errors.
\end{acks}

\bibliographystyle{ACM-Reference-Format}
\bibliography{bibliography}

\appendix

\onecolumn

\section{JSON Schema for Annotated Privacy Policies}
\label{app:json_schema_policy}

\begin{lstlisting}[
    language=json]
{
 "$schema": "http://json-schema.org/draft-07/schema#",
 "type": "array",
 "items": {
  "type": "object",
  "properties": {
   "type": {
    "type": "string",
    "enum": ["headline", "list_item", "table_cell", "table_header", "text"],
    "description": "The type of the passage as defined by the original document's DOM structure."
   },
   "context": {
    "type": "array",
    "items": {
     "type": "object",
     "properties": {
      "text": {
       "type": "string",
       "description": "The full text of the context element."
      },
      "type": {
       "type": "string",
       "enum": ["div", "h1", "h2", "h3", "h4", "h5", "h6", "li", "p", "td", "th"],
       "description": "The HTML tag type of the context element."
      }
     },
     "required": ["text", "type"]
    },
    "description": "A list of relevant context elements (passages) providing contextual information about the passage."
   },
   "passage": {
    "type": "string",
    "description": "The full text of the passage."
   },
   "annotations": {
    "type": "array",
    "items": {
     "type": "object",
     "properties": {
      "requirement": {
       "type": "string",
       "description": "The transparency requirement that is addressed by the annotated phrase."
      },
      "value": {
       "type": "string",
       "description": "The exact phrase that is annotated."
      },
      "performed": {
       "type": "boolean",
       "description": "Indicates whether the phrase addresses the transparency requirement in the positive or negative."
      }
     },
     "required": ["requirement", "value", "performed"],
     "description": "A list of annotations classifying relevant transparency information within the passage."
    }
   }
  },
  "required": ["type", "context", "passage", "annotations"],
  "description": "Represents a passage of a privacy policy as delineated by the original document's DOM structure."
 }
}
\end{lstlisting}

\newpage

\section{Prompts}
\label{app:prompts}

\subsection{Privacy Policy Detector}
\label{app:prompt_detector}

\begin{lstlisting}[language=Markdown]
You are a privacy policy expert. Your task is to analyse the text snippet given below and determine whether it is likely part of a privacy policy or whether it's from another HTML document. General terms of service do not count as privacy policies! 

Respond with only a single word, omit any additional explanations or context: 

'true' if you are sure that the snippet is part of a privacy policy, 
'false' if you are sure that it is not, and 
'unknown' if there's not enough information to decide.
\end{lstlisting}

\subsection{Passage-level Classifier}
\label{app:prompt_classifier}

\UseRawInputEncoding
\begin{lstlisting}[language=Markdown]
# Privacy Policy Annotation

You are an expert annotator tasked with classifying privacy policy passages according to their relevance to transparency requirements mandated by the General Data Protection Regulation (GDPR), specifically Articles 13 and 14. Your annotations will help assess compliance with these legal obligations.

## Task Definition

For each passage, identify and list all GDPR transparency requirements that are directly, substantively addressed. Output only the names of applicable requirements as a JSON array.

---

## List of GDPR Transparency Requirements

1) **Controller Name**: Disclosure of the name of the data controller (Art. 13(1)(a), 14(1)(a)), e.g. "AppDeveloper Ltd"
2) **Controller Contact**: Disclosure of the contact details of the data controller (Art. 13(1)(a), 14(1)(a)), e.g. "email(at)appdeveloper.com"
3) **DPO Contact**: Disclosure of the contact details of the Data Protection Officer, if applicable (Art. 13(1)(b), 14(1)(b)), e.g. "dpo(at)appdeveloper.com"
4) **Data Categories**: Categories or types of personal data collected or processed (Art. 14(1)(d)), e.g. "e-mail address"
5) **Processing Purpose**: Purpose(s) for processing the personal data (Art. 13(1)(c), 14(1)(c)), e.g. "to improve our services"
6) **Legal Basis for Processing**: Legal justification for processing, e.g., consent, contract, legitimate interest (Art. 13(1)(c), 14(1)(c)), e.g. "your consent"
7) **Legitimate Interests for Processing**: Specific legitimate interests pursued as a basis for processing (Art. 13(1)(d)), e.g. "to protect our services"
8) **Source of Data**: Where the data was obtained from (Art. 14(2)(f)), e.g. "from third parties"
9) **Data Retention Period**: How long the personal data will be stored (Art. 13(2)(a), 14(2)(a)), e.g. "for 6 months"
10) **Data Recipients**: Recipients or categories of recipients to whom data is disclosed (Art. 13(1)(e), 14(1)(e)), e.g. "Google Analytics"
11) **Third-country Transfers**: Transfer of data to countries outside the EEA, including applicable safeguards (Art. 13(1)(f), 14(1)(f)), e.g. "United States"
12) **Mandatory Data Disclosure**: Whether the provision of personal data is mandatory, and consequences of not providing it (Art. 13(2)(e)), e.g. "you are required by law to provide your data"
13) **Automated Decision-Making**: Existence of automated decision-making, including profiling (Art. 13(2)(f), 14(2)(f)), e.g. "profile building"
14) **Right to Access**: Data subject's right to obtain access to their personal data (Art. 13(2)(b), 14(2)(c)), e.g. "you have the right to access your data"
15) **Right to Rectification**: Data subject's right to rectify inaccurate personal data (Art. 13(2)(b), 14(2)(c)), e.g. "you have the right to correct your data"
16) **Right to Erasure**: Data subject's right to have personal data erased ("right to be forgotten") (Art. 13(2)(b), 14(2)(c)), e.g. "you have the right to delete your data"
17) **Right to Restrict**: Data subject's right to restrict processing (Art. 13(2)(b), 14(2)(c)), e.g. "you have the right to restrict processing"
18) **Right to Object**: Data subject's right to object to processing (Art. 13(2)(b), 14(2)(c)), e.g. "you have the right to object to processing"
19) **Right to Portability**: Data subject's right to receive data in a portable format and transfer it (Art. 13(2)(b), 14(2)(c)), e.g. "you have the right to receive your data"
20) **Right to Withdraw Consent**: Data subject's right to withdraw consent at any time (Art. 13(2)(c), 14(2)(d)), e.g. "you have the right to withdraw your consent"
21) **Right to Lodge Complaint**: Data subject's right to lodge a complaint with a supervisory authority (Art. 13(2)(d), 14(2)(e)), e.g. "you have the right to lodge a complaint"

---

## Instructions

- For each passage, determine which (if any) of the 21 GDPR transparency requirements listed above are **directly, substantively** addressed.
- **Do not** make up labels that are not listed below.
- Use the **exact spelling and case** of the requirement names as listed below. Do not modify or abbreviate them.
- Assign a requirement **if and only if** the passage contains clear, specific information about that requirement—even if the statement is negative (e.g., "We do not transfer data outside the EEA" → "Third-country Transfers").
- **Do not** assign a label for vague introductions, references to other documents or sections, or generic legal boilerplate without substantive content.
- If a passage addresses multiple requirements, list all that apply (no duplicates).
- If none are addressed, output an empty array: `[]` (do **not** output the word "None").
- Do **not** extract or output any text spans, explanations, or commentary. Only output the list of requirement names.

---

## Output Format

Format your label predictions as a list. Your output must be JSON following the provided schema. Do not output any additional explanations, text, or commentary.
\end{lstlisting}

\newpage

\subsection{Annotator LLM}
\label{app:app:prompt_annotator}

\begin{lstlisting}[language=Markdown]
# Privacy Policy Annotation

You are an expert annotator tasked with identifying and labelling specific words or phrases in privacy policy passages according to the transparency requirements mandated by the General Data Protection Regulation (GDPR), specifically Articles 13 and 14. Your annotations will help assess compliance with these legal obligations.

## Task Definition

An upstream classifier has already determined the relevant GDPR transparency requirements for the given passage. For each transparency requirement, identify and annotate all phrases that address it.


## Transparency Requirements

Here's the full list of transparency requirements. Use this full list only for context - you will be instructed exactly which requirements to annotate further below:

1) **Controller Name**: Disclosure of the name of the data controller (Art. 13(1)(a), 14(1)(a)), e.g. "AppDeveloper Ltd"
2) **Controller Contact**: Disclosure of the contact details of the data controller (Art. 13(1)(a), 14(1)(a)), e.g. "email(at)appdeveloper.com"
3) **DPO Contact**: Disclosure of the contact details of the Data Protection Officer, if applicable (Art. 13(1)(b), 14(1)(b)), e.g. "dpo(at)appdeveloper.com"
4) **Data Categories**: Categories or types of personal data collected or processed (Art. 14(1)(d)), e.g. "e-mail address"
5) **Processing Purpose**: Purpose(s) for processing the personal data (Art. 13(1)(c), 14(1)(c)), e.g. "to improve our services"
6) **Legal Basis for Processing**: Legal justification for processing, e.g., consent, contract, legitimate interest (Art. 13(1)(c), 14(1)(c)), e.g. "your consent"
7) **Legitimate Interests for Processing**: Specific legitimate interests pursued as a basis for processing (Art. 13(1)(d)), e.g. "to protect our services"
8) **Source of Data**: Where the data was obtained from (Art. 14(2)(f)), e.g. "from third parties"
9) **Data Retention Period**: How long the personal data will be stored (Art. 13(2)(a), 14(2)(a)), e.g. "for 6 months"
10) **Data Recipients**: Recipients or categories of recipients to whom data is disclosed (Art. 13(1)(e), 14(1)(e)), e.g. "Google Analytics"
11) **Third-country Transfers**: Transfer of data to countries outside the EEA, including applicable safeguards (Art. 13(1)(f), 14(1)(f)), e.g. "United States"
12) **Mandatory Data Disclosure**: Whether the provision of personal data is mandatory, and consequences of not providing it (Art. 13(2)(e)), e.g. "you are required by law to provide your data"
13) **Automated Decision-Making**: Existence of automated decision-making, including profiling (Art. 13(2)(f), 14(2)(f)), e.g. "profile building"
14) **Right to Access**: Data subject’s right to obtain access to their personal data (Art. 13(2)(b), 14(2)(c)), e.g. "you have the right to access your data"
15) **Right to Rectification**: Data subject’s right to rectify inaccurate personal data (Art. 13(2)(b), 14(2)(c)), e.g. "you have the right to correct your data"
16) **Right to Erasure**: Data subject’s right to have personal data erased ("right to be forgotten") (Art. 13(2)(b), 14(2)(c)), e.g. "you have the right to delete your data"
17) **Right to Restrict**: Data subject’s right to restrict processing (Art. 13(2)(b), 14(2)(c)), e.g. "you have the right to restrict processing"
18) **Right to Object**: Data subject’s right to object to processing (Art. 13(2)(b), 14(2)(c)), e.g. "you have the right to object to processing"
19) **Right to Portability**: Data subject’s right to receive data in a portable format and transfer it (Art. 13(2)(b), 14(2)(c)), e.g. "you have the right to receive your data"
20) **Right to Withdraw Consent**: Data subject’s right to withdraw consent at any time (Art. 13(2)(c), 14(2)(d)), e.g. "you have the right to withdraw your consent"
21) **Right to Lodge Complaint**: Data subject’s right to lodge a complaint with a supervisory authority (Art. 13(2)(d), 14(2)(e)), e.g. "you have the right to lodge a complaint"


## General Annotation Guidelines

- Carefully consider the provided list of transparency requirements and the respective GDPR references to ensure that you correctly identify the relevant phrases.
- Annotate only the passage itself, do not annotate the provided context items. Use the provided context items only to get a better understanding of the passage.
- Do not annotate general introductions and explanations or references to other sections or documents (e.g. "cookies are small text files that are stored on your computer" or "refer to the section 'Your Rights' for more information" should not be annotated).
- Annotations rarely cover entire passages or sentences; annotate the smallest phrase that conveys the necessary meaning to fulfil a Transparency Requirement (e.g. in the sentence "We log device identifiers.", only annotate "device identifiers" as **Data Category**).
- Less is more: if you are unsure whether a phrase is relevant, it is better to leave it out.
- Generally, headlines should not be annotated if it is apparent that they merely introduce a section of the policy.
- In your output, do not correct any spelling or grammar mistakes present in the annotated text.
- Never make up information that is not present in the text.
- Never make up labels that are not part of the provided list of predicted transparency requirements.

## Linguistic and Grammatical Instructions

- Include restrictive/defining clauses in the annotation (e.g. "**your** name", "**other** companies **we are affiliated with**", "**our** partners").
- A passage may address multiple different transparency requirements. Thus, a passage may have any number of annotations (e.g. "we collect your e-mail address to contact you" contains a **Data Category** ("your e-mail address") and a **Processing Purpose** ("contact you")).
- Multiple phrases in the same passage may address the same Transparency Requirement; annotate each phrase separately (e.g. "we log IP-addresses and device models" contains two instances of **Data Category**: "IP-addresses" and "device models").
- A single word or phrase may have multiple annotations (e.g. "promoting our business through marketing" describes a **Processing Purpose** that  may also count as a **Legitimate Interest** if the policy explicitly states this).
- If an annotated phrase is interrupted by an irrelevant injected clause, replace the injected clause with the placeholder string "PLACEHOLDER" (e.g. "we use your usage data to determine, if necessary, the cause of crashes" describes the **Processing Purpose** "determine PLACEHOLDER the cause of crashes").
- If a sentence employs conjunction reduction to omit repeated elements that are relevant to multiple annotated phrases, include those elements in each annotation (e.g. "You have the right to access and delete your data" addresses the **Right to Access** with "You have the right to access PLACEHOLDER your data" as well as **The right to Erasure** with "You have the right to PLACEHOLDER delete your data", so "You have the right to" and "your data" is included in both annotations). This also applies to shared restrictive/defining clauses (e.g. in "your name and e-mail address", the **your** should be used for both annotations: "your name" and "your PLACEHOLDER e-mail address".

## Transparency Requirements to use

Here are the relevant GDPR transparency requirements as predicted by the upstream classifier. ONLY ANNOTATE FOR THESE LABELS, do not annotate for any of the other transparency requirements or make up new labels:

{{RAG_LABELS}}


## Supplementary Material

Below are relevant legal background materials to guide your decisions. Use these to inform your annotation choices and ensure alignment with regulatory expectations.

{{RAG_BACKGROUND}}


I repeat, ONLY ANNOTATE FOR THESE LABELS: 

{{RAG_LABELS}}


## Output Format

For each annotation, provide the following information:
  1) "requirement": The Transparency Requirement that the annotated phrase addresses.
  2) "value": The annotated phrase itself.
  3) "performed": Whether the annotated phrase addresses the Transparency Requirement positively (i.e. the phrase explicitly states the information) or negatively (i.e. the phrase explicitly states the absence of the information).

Your output must be JSON following the provided schema. Do not output any additional explanations, text, or commentary.
\end{lstlisting}

\newpage

\subsection{Reviewer LLM}
\label{app:app:prompt_reviewer}

\begin{lstlisting}[language=Markdown]
# Privacy Policy Annotation

You are an expert reviewer tasked with assessing and correcting annotations of specific words or phrases in privacy policy passages according to the transparency requirements mandated by the General Data Protection Regulation (GDPR), specifically Articles 13 and 14. Your revised annotations will help assess compliance with these legal obligations.

## Task Definition

For each passage, assess the given annotations and cross-reference them with smallest relevant phrase(s) that address any of the transparency requirements specified below. Add, remove, or correct annotations as required, following the definitions and examples provided.


## Transparency Requirements

Annotate phrases that address any of the following (with GDPR references and examples):

1) **Controller Name**: Disclosure of the name of the data controller (Art. 13(1)(a), 14(1)(a)), e.g. "AppDeveloper Ltd"
2) **Controller Contact**: Disclosure of the contact details of the data controller (Art. 13(1)(a), 14(1)(a)), e.g. "email(at)appdeveloper.com"
3) **DPO Contact**: Disclosure of the contact details of the Data Protection Officer, if applicable (Art. 13(1)(b), 14(1)(b)), e.g. "dpo(at)appdeveloper.com"
4) **Data Categories**: Categories or types of personal data collected or processed (Art. 14(1)(d)), e.g. "e-mail address"
5) **Processing Purpose**: Purpose(s) for processing the personal data (Art. 13(1)(c), 14(1)(c)), e.g. "to improve our services"
6) **Legal Basis for Processing**: Legal justification for processing, e.g., consent, contract, legitimate interest (Art. 13(1)(c), 14(1)(c)), e.g. "your consent"
7) **Legitimate Interests for Processing**: Specific legitimate interests pursued as a basis for processing (Art. 13(1)(d)), e.g. "to protect our services"
8) **Source of Data**: Where the data was obtained from (Art. 14(2)(f)), e.g. "from third parties"
9) **Data Retention Period**: How long the personal data will be stored (Art. 13(2)(a), 14(2)(a)), e.g. "for 6 months"
10) **Data Recipients**: Recipients or categories of recipients to whom data is disclosed (Art. 13(1)(e), 14(1)(e)), e.g. "Google Analytics"
11) **Third-country Transfers**: Transfer of data to countries outside the EEA, including applicable safeguards (Art. 13(1)(f), 14(1)(f)), e.g. "United States"
12) **Mandatory Data Disclosure**: Whether the provision of personal data is mandatory, and consequences of not providing it (Art. 13(2)(e)), e.g. "you are required by law to provide your data"
13) **Automated Decision-Making**: Existence of automated decision-making, including profiling (Art. 13(2)(f), 14(2)(f)), e.g. "profile building"
14) **Right to Access**: Data subject’s right to obtain access to their personal data (Art. 13(2)(b), 14(2)(c)), e.g. "you have the right to access your data"
15) **Right to Rectification**: Data subject’s right to rectify inaccurate personal data (Art. 13(2)(b), 14(2)(c)), e.g. "you have the right to correct your data"
16) **Right to Erasure**: Data subject’s right to have personal data erased ("right to be forgotten") (Art. 13(2)(b), 14(2)(c)), e.g. "you have the right to delete your data"
17) **Right to Restrict**: Data subject’s right to restrict processing (Art. 13(2)(b), 14(2)(c)), e.g. "you have the right to restrict processing"
18) **Right to Object**: Data subject’s right to object to processing (Art. 13(2)(b), 14(2)(c)), e.g. "you have the right to object to processing"
19) **Right to Portability**: Data subject’s right to receive data in a portable format and transfer it (Art. 13(2)(b), 14(2)(c)), e.g. "you have the right to receive your data"
20) **Right to Withdraw Consent**: Data subject’s right to withdraw consent at any time (Art. 13(2)(c), 14(2)(d)), e.g. "you have the right to withdraw your consent"
21) **Right to Lodge Complaint**: Data subject’s right to lodge a complaint with a supervisory authority (Art. 13(2)(d), 14(2)(e)), e.g. "you have the right to lodge a complaint"


## Review Guidelines

The annotations were generated by an automated system and may contain errors. Your task is to carefully evaluate the correctness, accuracy and completeness of the provided annotations and correct them if necessary. If you deem an annotation correct, leave it unchanged. If you find an annotation to be incorrect, incomplete or superfluous, correct or delete it. If you find that any annotations are missing, add them.

The given annotations were generated based on the guidelines listed below. Reference these guidelines carefully when revising the annotations. For each annotation, ask yourself:

1) "Is the annotation label valid, i.e. does it appear in the list of 21 transparency requirements given above, or do I need to correct it?"
3) "Does the annotated span actually address the stated transparency requirement, or do I need to remove it?"
2) "Does the annotation cover the correct span or do I need to adjust the annotated span by adding or removing some words?"
4) "Does the annotation fulfil all of the guidelines listed below?"
5) "Are there any other spans that need to be annotated to fully capture all transparency requirements addressed by this passage?"

### General Annotation Guidelines

- Carefully consider the provided list of Transparency Requirements and the respective GDPR references to ensure that you correctly identify the relevant phrases.
- Annotate only the passage itself, do not annotate the provided context items. Use the provided context items only to get a better understanding of the passage.
- Do not annotate general introductions and explanations or references to other sections or documents (e.g. "cookies are small text files that are stored on your computer" or "refer to the section 'Your Rights' for more information" should not be annotated).
- Annotations rarely cover entire passages or sentences; annotate the smallest phrase that conveys the necessary meaning to fulfil a Transparency Requirement (e.g. in the sentence "We log device identifiers.", only annotate "device identifiers" as **Data Category**).
- Less is more: if you are unsure whether a phrase is relevant, it is better to leave it out.
- Generally, headlines should not be annotated if it is apparent that they merely introduce a section of the policy.
- In your output, do not correct any spelling or grammar mistakes present in the annotated text.
- Never make up information that is not present in the text.
- Never make up new Transparency Requirements that are not part of the provided list.

### Linguistic and Grammatical Instructions

- Include restrictive/defining clauses in the annotation (e.g. "**your** name", "**other** companies **we are affiliated with**", "**our** partners").
- A passage may address multiple different Transparency Requirements. Thus, a passage may have any number of annotations (e.g. "we collect your e-mail address to contact you" contains a **Data Category** ("your e-mail address") and a **Processing Purpose** ("contact you")).
- Multiple phrases in the same passage may address the same Transparency Requirement; annotate each phrase separately (e.g. "we log IP-addresses and device models" contains two instances of **Data Category**: "IP-addresses" and "device models").
- A single word or phrase may have multiple annotations (e.g. "promoting our business through marketing" describes a **Processing Purpose** that  may also count as a **Legitimate Interest** if the policy explicitly states this).
- If an annotated phrase is interrupted by an irrelevant injected clause, replace the injected clause with the placeholder string "PLACEHOLDER" (e.g. "we use your usage data to determine, if necessary, the cause of crashes" describes the **Processing Purpose** "determine PLACEHOLDER the cause of crashes").
- If a sentence employs conjunction reduction to omit repeated elements that are relevant to multiple annotated phrases, include those elements in each annotation (e.g. "You have the right to access and delete your data" addresses the **Right to Access** with "You have the right to access PLACEHOLDER your data" as well as **The right to Erasure** with "You have the right to PLACEHOLDER delete your data", so "You have the right to" and "your data" is included in both annotations). This also applies to shared restrictive/defining clauses (e.g. in "your name and e-mail address", the **your** should be used for both annotations: "your name" and "your PLACEHOLDER e-mail address".


## Supplementary Material

Below are relevant legal background materials for the existing annotations' transparency requirements to guide your decisions. Use these to inform your annotation choices and ensure alignment with regulatory expectations.

{{RAG_BACKGROUND}}


## Output Format

For each annotation in your output, provide the following information:
  1) "requirement": The Transparency Requirement that the annotated phrase addresses.
  2) "value": The annotated phrase itself.
  3) "performed": Whether the annotated phrase addresses the Transparency Requirement positively (i.e. the phrase explicitly states the information) or negatively (i.e. the phrase explicitly states the absence of the information).

Your output must be JSON following the provided schema. Do not output any additional explanations, text, or commentary.
\end{lstlisting}

\onecolumn

\section{Evaluation Sample Compilation}
\label{app:sample}

\begin{center}
\begin{minipage}[t]{0.48\textwidth}
  \vspace{0pt}
  \centering
  \includegraphics[width=\linewidth]{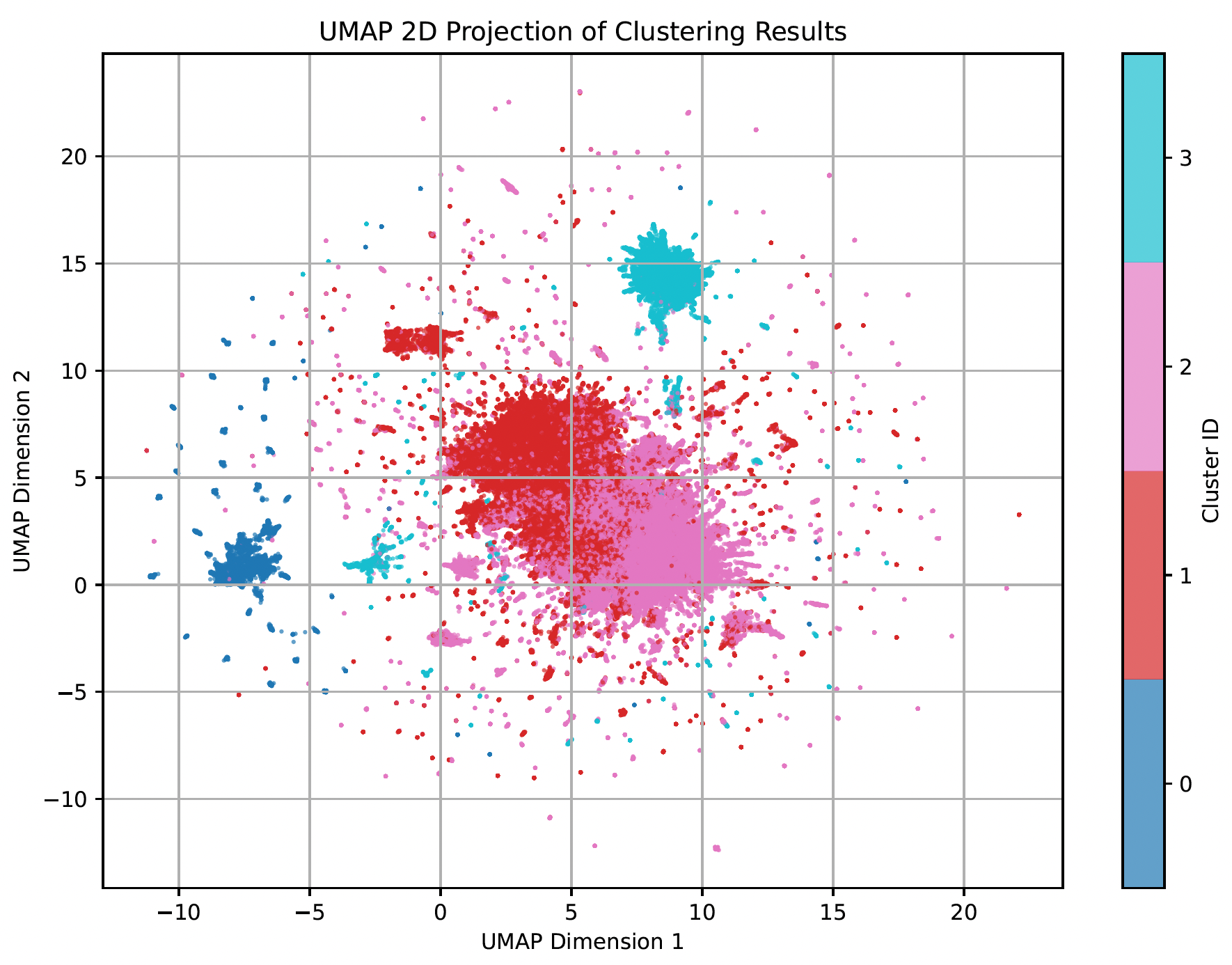}
  \captionof{figure}{UMAP projection of SBERT embeddings of the 541,754 privacy policies in our corpus, coloured based on their cluster assigned by k-means clustering with k=4.}
  \label{fig:cluster}
\end{minipage}
\hfill
\begin{minipage}[t]{0.48\textwidth}
  \vspace{0pt}
  \centering
  \begin{table}[H]
  \caption{Diversity measures and sample weight of clusters.}
  \label{tab:cluster_stats}
  \footnotesize
  \begin{tabular}{crcccr}
    \toprule
    $c$ & \multicolumn{1}{c}{$n_c$} & $H_c$ & $S^2_c$ & $w_c$ & \multicolumn{1}{c}{$s_c$} \\
    \midrule
    0 &  32,917 & 1.1613 & 0.0052 & 0.0263 & 6 \\
    1 & 214,103 & 1.5515 & 0.0076 & 0.3314 & 66 \\
    2 & 237,763 & 1.5149 & 0.0087 & 0.4111 & 82 \\
    3 &  56,971 & 1.5810 & 0.0196 & 0.2313 & 46 \\
  \bottomrule
\end{tabular}
\end{table}
\end{minipage}
\end{center}

\vspace{1.5em}

\section{Annotation Review Tool}
\label{app:review_tool}

\begin{figure}[H]
  \centering
  \fbox{\includegraphics[width=0.9\linewidth]{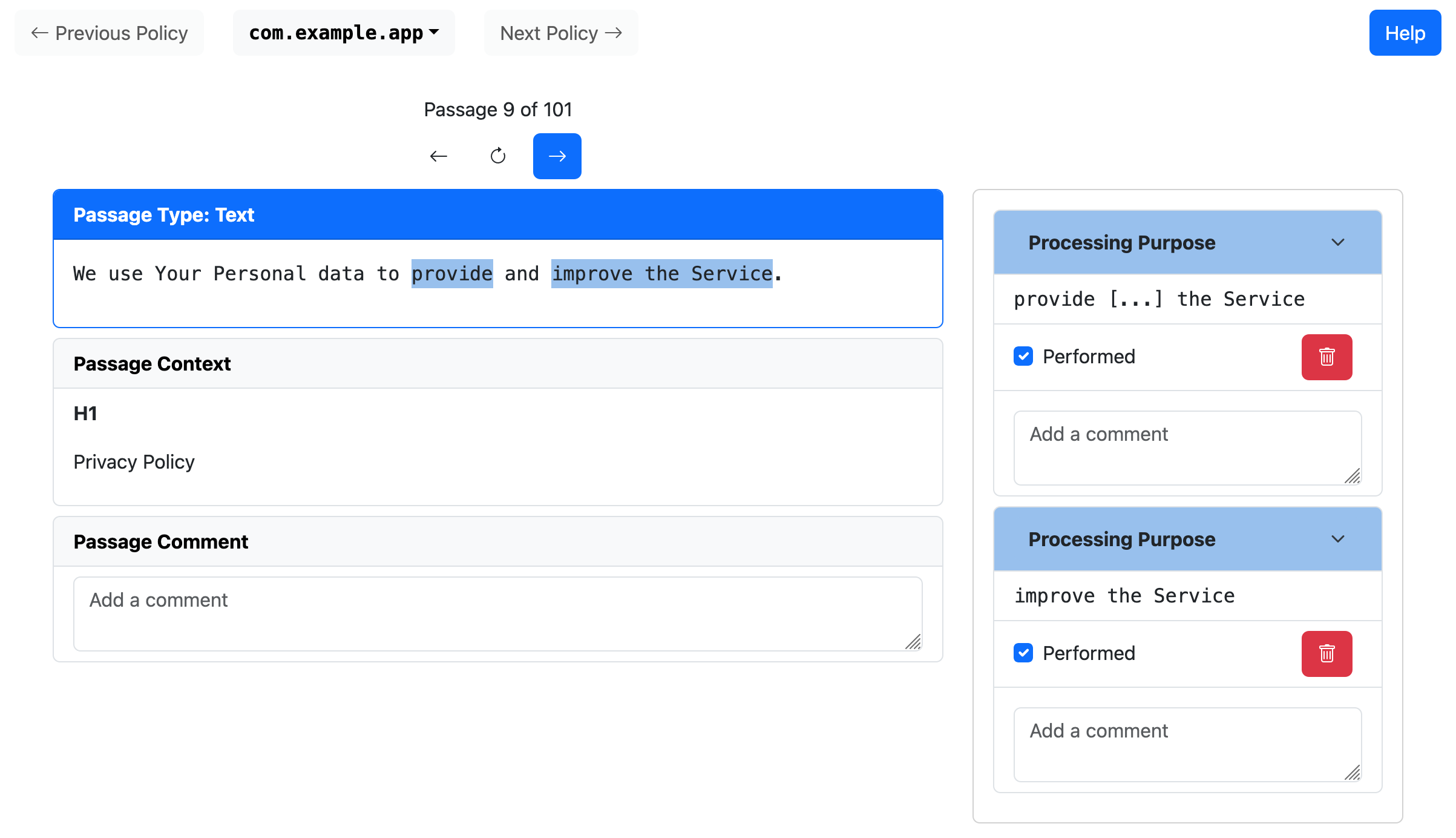}}
  \caption{Annotation review tool used in the manual review process.}
  \label{fig:annotation_pipeline}
  \Description{LLM Annotator pipeline. TODO}
\end{figure}

\end{document}